\documentclass{bmvc2k}

\title{K-CLUE: A Segmentation-to-Clustering Framework for Ki67 Cell Typing}

\addauthor{Ana Beatriz Vieira}{anabeatrizvieira@tecnico.ulisboa.pt}{1,2}
\addauthor{Sara P. Oliveira}{s.oliveira@nki.nl}{3}
\addauthor{Diana Montezuma}{diana.felizardo@impdiagnostics.com}{4}
\addauthor{João Fraga}{joao.fraga@impdiagnostics.com}{4}
\addauthor{Jaime S. Cardoso}{jaime.cardoso@fe.up.pt}{5,6}
\addauthor{Arlindo L. Oliveira}{arlindo.oliveira@tecnico.ulisboa.pt}{1,2}

\addinstitution{
 Instituto Superior Técnico,\\
 Universidade de Lisboa, Portugal\\
}
\addinstitution{
 INESC-ID, Lisboa, Portugal\\
}

\addinstitution{
 The Netherlands Cancer Institute\\
 Amsterdam, The Netherlands\\
}

\addinstitution{
IMP Diagnostics, Porto, Portugal\\
}

\addinstitution{
 Faculdade de Engenharia,\\
 Universidade do Porto, Portugal\\
}

\addinstitution{
 INESC-TEC, Porto, Portugal\\
}

\runninghead{VIEIRA ET AL.}{KI67 CELL TYPING}

\def\etal{\emph{et al}\bmvaOneDot}

\usepackage{booktabs}
\usepackage{multirow}
\usepackage[table]{xcolor}
\usepackage{wrapfig}

\begin{document}

\maketitle

\begin{abstract}
Ki67 is a key marker of tumor proliferation in breast cancer, but its quantification from immunohistochemistry images normally depends on supervised classifiers trained per dataset. We examine the role of frozen pretrained visual representations in annotation-free Ki67 cell typing. To this end, we use K-CLUE, a framework that segments individual nuclei, represents each nucleus with a pretrained encoder, reduces the resulting embeddings with principal component analysis, and separates Ki67-positive from Ki67-negative nuclei using $k$-means clustering, without task-specific training or manual annotations. Clustering the raw components is highly sensitive to the choice of encoder: across seven encoders, the Ki67-positive F1-score on matched MIST nuclei varies by 44 points, although Ki67 status remains linearly decodable from every representation, with supervised probe AUCs above 0.98. Analysis of the embedding spaces shows that clustering succeeds when DAB-related variation occupies the leading directions of variance. Whitening the retained principal components removes this dependence, reducing the spread across encoders to 1.00 F1-score point. Handcrafted color representations are also competitive, confirming that Ki67 status is strongly reflected in chromatic information, but they require an explicit choice of color space. In end-to-end evaluation, K-CLUE reaches F1-scores of 85.00\% and 81.91\% for Ki67-positive and Ki67-negative nuclei on MIST, and 79.59\% and 81.16\% on BCData, outperforming DeepLIIF and IHCScoreGAN on both datasets. These results show that pretrained representations can support annotation-free Ki67 typing when their variance structure is accounted for.
\end{abstract}

%-------------------------------------------------------------------------
\section{Introduction}
Breast cancer remains one of the most critical challenges in global healthcare and is currently the most frequently diagnosed malignancy among women worldwide. Despite substantial therapeutic advances, early and accurate tumor characterization remains essential, as many cases are still diagnosed at advanced stages, limiting treatment efficacy~\cite{Santos2025}. 

Modern oncology has therefore shifted toward precision medicine, where treatment decisions are tailored to the molecular profile of the tumor. In breast cancer, stratification relies on immunohistochemistry (IHC) biomarkers, such as estrogen receptor (ER), progesterone receptor (PgR), and HER2. While these markers capture key hormonal and genetic dependencies, they provide limited information regarding tumor proliferative activity and aggressiveness~\cite{AlTarawneh2025}.

The Ki67 biomarker addresses this limitation. Ki67 is a nuclear protein expressed during the active phases of the cell cycle ($G_1, S, G_2,$ and mitosis) and absent in resting cells ($G_0$)~\cite{Santos2025,Davey2021,Aung2021}. In practice, Ki67 is visualized using a dual H-DAB staining protocol that colors proliferating nuclei brown and all other nuclei blue. Accurate classification of individual nuclei as Ki67-positive or Ki67-negative is therefore the critical step underlying assessment of proliferative activity, subtype stratification, and chemotherapy planning.

Despite its clinical relevance, Ki67 quantification remains largely manual. Whole-slide images (WSIs) may contain hundreds of thousands of nuclei, making exhaustive counting impractical. Pathologists, therefore, estimate the index by visually inspecting selected regions of interest (ROIs), typically high-proliferation areas (hotspots) at the invasive front of the tumor, and assessing the proportion of positive and negative nuclei. Since any detectable brown nuclear staining is considered positive, independent of staining intensity, borderline cases require subjective interpretation, leading to significant inter-observer variability~\cite{Dy2024}. 

Given the clinical importance of reliable Ki67 assessment, there is a clear need for automated and reproducible computational methods that do not depend on costly nuclear annotations. To address this, we introduce K-CLUE, an annotation-free pipeline for Ki67 cell typing. Individual nuclei are first segmented, each nuclear crop is represented using a frozen pretrained encoder, the embeddings are reduced with principal component analysis (PCA), and the retained components are clustered into Ki67-positive and Ki67-negative groups.

The choice of pretrained encoder, however, has a substantial effect on the resulting clusters. Keeping the segmentation, nuclear crops, and clustering procedure fixed, changing only the encoder causes the Ki67-positive F1-score to vary by more than forty percentage points. This variation does not follow encoder capacity or pretraining domain.

We show that these differences arise from the geometry of the embedding space rather than from the absence of Ki67-relevant information. Since $k$-means operates on Euclidean distances, directions of high variance have greater influence on the resulting clustering. Ki67 status is therefore recovered when staining-related variation is prominent among the leading principal components, but may be ignored when other sources of variation dominate them. A supervised linear probe confirms that Ki67 status is linearly decodable from every encoder we evaluate, including those whose unsupervised clustering performance is close to chance. Whitening the retained principal components reduces this dependence by rescaling them to unit variance and preventing individual high-variance directions from dominating the clustering objective.

We further compare these learned representations with handcrafted color descriptors. Their strong performance confirms that Ki67 typing is largely driven by chromatic information, while also showing that the most effective handcrafted color representation varies across datasets. K-CLUE instead recovers the relevant separation from generic frozen representations without requiring a task-specific color descriptor. The contributions of this paper are the following:
\begin{itemize}
    \item We show that whitening the principal components makes the pipeline largely independent of the encoder, removing the need to select a representation or color descriptor per dataset.
    
    \item We validate the resulting pipeline, K-CLUE, end-to-end on two public cohorts under a single configuration, against supervised, generative, and handcrafted color baselines.
\end{itemize}

Annotations are used only for evaluation and for the analyses explicitly identified as supervised probes, but never by the method at inference.

\section{Related Work}
\paragraph{Biomarker Ki67:}
%Geread \etal~\cite{10.3389/fbioe.2019.00226} proposed an unsupervised color separation framework for Ki67 proliferation index calculation based on IHC color histograms (IHCCH), isolating the diaminobenzidine and hematoxylin stains without expert annotations. An adaptive overstaining threshold and a nuclei radius estimator mitigate background noise and scale variability, and the method reaches $92.5\%$ accuracy on Protein Atlas samples while remaining consistent across staining protocols and hardware. 
Geread \etal~\cite{10.3389/fbioe.2019.00226} proposed an unsupervised color-separation framework based on IHC color histograms (IHCCH) for Ki67 quantification. The method separates DAB and haematoxylin staining without expert annotations and includes an overstaining threshold and an automated nuclei-radius estimator to improve robustness to staining variability and differences in nuclear scale. On 30 TMA cores with nucleus-level ground truth, the method achieved an average proliferation-index difference of 3.25\%. Generalization was also evaluated on 50 Protein Atlas TMAs, for which only PI ranges were available. Across both datasets, 74 of 80 images were assigned to the correct PI range.
 
Generalizable cellular segmentation has advanced considerably in recent years. Pachitariu \etal~\cite{Pachitariu2025} introduced Cellpose-SAM, which replaces the Cellpose backbone with the Segment Anything Model~\cite{Kirillov_2023_ICCV} and approaches a human-consensus reference for nucleus segmentation across imaging conditions, enabling deployment on heterogeneous datasets with minimal retraining. Other recent architectures include the transformer-based CellViT++~\cite{cellvit}, the contour-aware HoVer-NeXt~\cite{baumann2024hovernext}, star-convex instance segmentation with StarDist~\cite{weigert2022}, and detection-based CellDETR~\cite{pmlr-v250-pina24a}. These models are effective on H\&E and fluorescence data but rely on appearance assumptions that transfer poorly to DAB-stained IHC, which limits their use as drop-in components without substantial retraining.
 
Two generative frameworks target IHC scoring directly. Ghahremani \etal~\cite{Ghahremani2022} introduced DeepLIIF, a multitask model that performs stain deconvolution and cell segmentation jointly, trained on co-registered immunohistochemistry and multiplex immunofluorescence (mpIF) images of the same tissue sections. A conditional generative adversarial network learns the mapping between the two modalities to synthesize protein-specific channels that serve as an objective reference for cell-level analysis, which avoids manual labeling at the cost of requiring paired acquisitions. Molnar \etal~\cite{ihcscoregan} removed that requirement with IHCScoreGAN, formulating Ki67 scoring as unpaired domain transfer and using a CycleGAN architecture with a dual-branch generator that produces synthetic masks, cell centers and biomarker expression maps from unpaired public H\&E data. Validated on an internal cohort of over 2100 cases and on the external BCData benchmark, the method matches or exceeds supervised baselines, which shows that unsupervised domain transfer is a scalable alternative to supervised and multimodal frameworks.

\paragraph{Whitening:}
Whitening is a well-established post-processing technique for learned representations. Jégou and Chum~\cite{10.1007/978-3-642-33709-3_55} applied PCA whitening to image-retrieval descriptors, showing that removing correlations induced by co-occurring visual features improves retrieval performance. In natural language processing, Su et al.~\cite{su2021} used whitening to reduce the anisotropy of sentence embeddings and obtain more uniformly distributed representations, while Huang et al.~\cite{huang-etal-2021-whiteningbert-easy} extended this idea to BERT-based sentence embeddings and reported consistent improvements across semantic textual similarity tasks. More recently, Saraswatula and Klindt~\cite{saraswatula2025} studied whitening for sparse autoencoders, showing that it improves the conditioning of the optimization problem by reducing the influence of correlated and high-variance directions.

\section{Methodology}
% This section describes the datasets, segmentation/detection models, and stain-aware strategies considered in our work, as well as the proposed unsupervised pipeline for separating Ki67-positive and Ki67-negative nuclei directly from IHC images.

\subsection{Datasets}
\begin{wrapfigure}[16]{r}{0.24\textwidth}
    \vspace{-\intextsep}
    \centering
    \includegraphics[width=\linewidth]{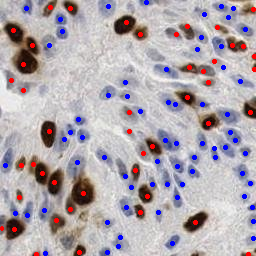}
    \vspace{-0.6cm}
    \caption{MIST tile.}
    \label{fig:mist_data}
    \vspace{0.2cm}
    \includegraphics[width=\linewidth]{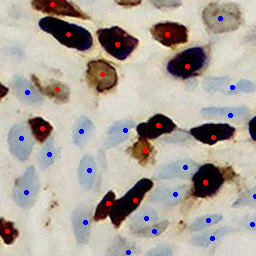}
    \vspace{-0.6cm}
    \caption{BCData tile.}
    \label{fig:bcdata_data}
    \vspace{-\intextsep}
\end{wrapfigure}

\paragraph{MIST:} We use a subset of the original Multi-IHC Stain Translation (MIST) dataset~\cite{li2023adaptive}, selecting one tile ($1024\times1024$ pixels, 20X magnification, resolution of 0.4661 µm/pixel) per WSI ID (n=56). To match the input size expected by the segmentation models, we randomly cropped a $256\times256$ tile from the original image. The selected tiles were manually annotated by a pathologist, with Ki67-positive (red) and Ki67-negative (blue) cells marked (point annotations) using distinct labels, as illustrated in Figure \ref{fig:mist_data}. \par

\paragraph{BCData:} The Breast tumor Cell Dataset (BCData)~\cite{10.1007/978-3-030-59722-1_28} consists of 1,338 ROIs ($640\times640$ pixels), extracted from 394 Ki67-stained breast WSIs (40X magnification, resolution of 0.2239 µm/pixel), with a total of 181,074 tumor cells annotated (point annotations) as Ki67-positive (62,623) and Ki67-negative (118,451) cells, as illustrated in Figure \ref{fig:bcdata_data}. Similarly to the MIST dataset, for model compatibility, we randomly cropped a $256\times256$ tile from the original images in the test set (n=402). \par

\subsection{Object segmentation}
Accurate cell detection is a prerequisite for robust Ki67 quantification, since subsequent analysis depends directly on the completeness and boundary quality of the detected nuclei. We therefore began by detecting all nuclei in each IHC image, without attempting to label them as Ki67 positive (brown) or negative (blue) at this stage. This design choice avoids introducing early bias from stain-dependent heuristics and ensures that downstream steps operate on a consistent set of candidate cells.

We considered several state-of-the-art nuclei segmentation and detection models, including Cellpose-SAM~\cite{Pachitariu2025}, HoVer-NeXt~\cite{baumann2024hovernext}, StarDist~\cite{weigert2022}, CellViT++~\cite{cellvit}, and CellDETR~\cite{pmlr-v250-pina24a}. We also considered Cellpose-DINOv3\footnote{\url{https://github.com/mouseland/cellpose}}, a variant distributed with the Cellpose codebase in which the SAM backbone is replaced by a DINOv3 encoder. In practice, many of these methods are predominantly trained and validated on H\&E datasets, where nuclear appearance differs substantially from IHC. Consequently, approaches that perform well on hematoxylin-dominant nuclei may under-segment or miss DAB-stained nuclei.

Based on these considerations, we adopt Cellpose-DINOv3 ViT-B as our nuclei segmentation model for all subsequent experiments, as it provides masks suitable for downstream processing across IHC images. A quantitative comparison of state-of-the-art nuclei segmentation and detection models is reported in Section~\ref{sec:seg_models}.

\subsection{K-CLUE}
K-CLUE is an annotation-free pipeline for separating Ki67-positive and Ki67-negative nuclei directly from RGB tiles, used throughout this work, both as the object of the analysis (Section \ref{sec:whitening}) and as the method evaluated (Section \ref{sec:results}). Given an input tile, we first apply Cellpose-DINOv3 in inference mode to obtain instance-level nuclear masks. Each detected nucleus is then isolated in a fixed $64\times64$ pixel bounding box centered on its centroid and upsampled to the $224\times224$ input resolution expected by the encoders. The bounding box size corresponds to the largest nucleus in the dataset plus one pixel of padding, so that every nucleus fits entirely inside the crop. To reduce interference from neighboring cells, pixels outside the nuclear mask are replaced with the median background color of the entire dataset. This avoids artificial high-contrast boundaries and encourages the encoder to focus on intrinsic chromogenic characteristics. Alternatives for the background fill and for the inclusion of neighboring cells in the crop are reported in Appendix~\ref{A.3}.

Each crop is then encoded by a frozen pretrained backbone. We consider seven encoders from three families, namely generic convolutional neural networks (ResNet-101~\cite{he2016deep}, EfficientNet-B2~\cite{pmlr-v97-tan19a}, ConvNeXt-V2 Tiny~\cite{10205236}), a self-supervised vision model (DINOv3 ViT-S/16~\cite{simeoni2025dinov3}), and pathology-specific encoders (HoVer-NeXt~\cite{baumann2024hovernext}, UNI2-h~\cite{chen2024uni}, Virchow2~\cite{zimmermann2024virchow2scalingselfsupervisedmixed}). No encoder is fine-tuned or adapted at any point, and all of them receive the same crops.

The resulting embeddings are projected via principal component analysis, retaining the components that account for 90\% of the variance. Each retained component is then rescaled to unit variance, an operation known as whitening. Whitening uses neither Ki67 annotations nor DAB measurements. We treat it as a configurable step rather than a fixed one, since Section~\ref{sec:whitening} compares the same pipeline with and without it.

\begin{figure}[ht]
    \centering
    \includegraphics[width=1\linewidth]{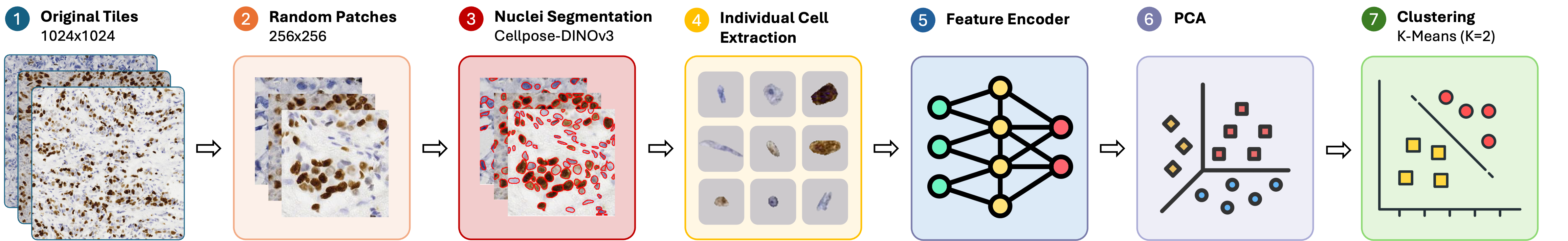}
    \caption{Overview of the proposed method pipeline. An input RGB tile is first segmented using Cellpose-DINOv3 to obtain instance-level nuclear masks. Each nucleus is extracted as a centered crop with background masking. Deep representations are computed with a pretrained feature encoder, reduced via PCA, whitened, and clustered using $k$-means to separate Ki67-positive from Ki67-negative nuclei.}
    \label{fig:our_method}
\end{figure}

Finally, $k$-means ($k=2$) is applied to the pooled embeddings from all dataset images jointly, yielding a single global decision boundary rather than a per-image one. This prevents images with a homogeneous cell population from being forced into a balanced split. Each cluster is then labeled according to its mean DAB intensity, with the higher-intensity cluster designated Ki67-positive and the other Ki67-negative. The DAB measurement therefore contributes a single binary decision and is not used to represent or group the nuclei. Since Cellpose-DINOv3 generates instance segmentation masks and each nucleus's instance ID is retained throughout the pipeline, predicted labels can be traced directly back to their corresponding instance in the original mask. Figure~\ref{fig:our_method} provides an overview of the full pipeline. 

\subsection{Analysis of the embedding space}
\label{sec:embeddings}
The pipeline described above combines two operations whose interaction is not obvious. The encoder determines what is represented, while $k$-means determines what is recovered from that representation, and the two need not agree. A representation can encode Ki67 status without the clustering separating nuclei accordingly.

The reason lies in how $k$-means measures similarity. After PCA, each nucleus is described by its position along a set of orthogonal directions, and the contribution of a direction to the distance between two nuclei grows with the variance along it. The resulting clusters therefore follow whichever attribute the highest-variance directions encode. Staining influences the clustering only if it is represented in those directions, not elsewhere in the embedding. When an unrelated attribute occupies the highest-variance directions, the clusters correspond to that attribute rather than Ki67 status. Whitening rescales every retained direction to unit variance, so that no attribute is favoured because it varies more.

Separating the two cases requires measuring them independently, and we report two quantities for each encoder. The first indicates whether Ki67 status is present in a representation. We fit a logistic regression classifier on the unwhitened embeddings, with cross-validation grouped by tile ID, and report the mean area under the ROC curve over the matched nuclei. This quantity uses ground-truth labels and is used for analysis only. The second indicates whether that information lies in the directions used by $k$-means. For each encoder we take the first principal component, which carries the largest variance and therefore has the greatest influence on the distance, and compute the absolute Pearson correlation between its projections and a set of per-nucleus covariates measured from the image: the mean DAB and hematoxylin optical densities, obtained by color deconvolution~\cite{ruifrok2001color}, the mean saturation and luminance, the nuclear area, and the tile of origin, for which we report the correlation ratio ($\eta$) since it is categorical. These covariates require no annotations. %DAB and luminance are close to collinear under H-DAB staining, so we also report the partial correlation with DAB when luminance is held fixed. These covariates require no annotations.

\section{Experiments}
\subsection{Data Preprocessing}
\label{sec:processing}
Although the proposed method does not require resolution adjustments, IHCScoreGAN expects $256\times256$ input images at 0.2239~\textmu m/pixel. Since the MIST dataset has a resolution of 0.4661~\textmu m/pixel, each $256\times256$ tile was upsampled $\sim$2X (to $512\times512$) to approximate the target resolution while preserving the original field of view, then divided into four non-overlapping $256\times256$ sub-tiles, independently processed by IHCScoreGAN to generate probabilistic heatmaps. To prevent double-counting at tile boundaries, the four heatmaps were stitched into a single $512\times512$ probability map, and cell detection was performed once on this stitched heatmap, with counts attributed back to the original tile. For BCData, the main challenge was input size, as IHCScoreGAN requires $256\times256$ images, while for DeepLIIF, which expects $256\times256$ at 20X or $512\times512$ at 40X, we adjusted the code to accommodate the BCData subset ($256\times256$ images at 40X magnification), ensuring the model's cell-size requirements were met.

\subsection{Evaluation protocol}
\label{sec:containment}
Since the ground-truth annotations are points rather than masks, predicted nuclei are matched to annotations using one-to-one Hungarian assignment between ground-truth points and predicted nuclear centroids. Distances are measured in physical units, and matches are accepted within a \(3\,\mu\mathrm{m}\) tolerance, corresponding approximately to the mean nuclear radius across MIST and BCData. Using a physical rather than pixel-based threshold provides a consistent matching criterion across datasets with different spatial resolutions.

We use two evaluation settings. For the representation-level analyses, including the encoder and handcrafted-feature comparisons, classification metrics are computed only over matched nuclei. This isolates the effect of the representation and clustering stages from upstream detection errors. In the final method comparisons reported in Section \ref{sec:results}, we instead use end-to-end evaluation. Matched nuclei with the correct class are counted as true positives, matched nuclei with an incorrect class contribute a false positive for the predicted class and a false negative for the ground-truth class, unmatched predictions are counted as false positives, and unmatched annotations as false negatives. Counting performance is evaluated independently from matching using the total number of predicted Ki67-positive and Ki67-negative nuclei in each image.

\subsection{Nuclei segmentation}
\label{sec:seg_models}
This section quantitatively compares representative nuclei segmentation and detection models on the MIST dataset to select a robust detector for all subsequent experiments. Since all downstream steps depend on the detected nuclei, the primary requirement here is robust detection across the staining variability present in IHC images, including both hematoxylin-dominant (blue) nuclei and DAB-stained (brown) nuclei.

All models were run on the same IHC tiles using the recommended inference configurations. For Cellpose-DINOv3 we use the ViT-B encoder, which outperformed the larger ViT-L variant on MIST at lower inference cost (Appendix~\ref{A.1}). Cellpose-DINOv3 ViT-B is referred to as Cellpose-DINOv3 hereafter.

All methods were evaluated with the procedure described in the previous section. Table~\ref{tab:tab1} summarizes the resulting detection performance, with precision, recall, and F1-Score computed from spatially matched detections, along with MAE, the mean absolute error between predicted and ground-truth nucleus counts, averaged across tiles.

\begin{figure}[ht]
    \centering
    \includegraphics[width=1\linewidth]{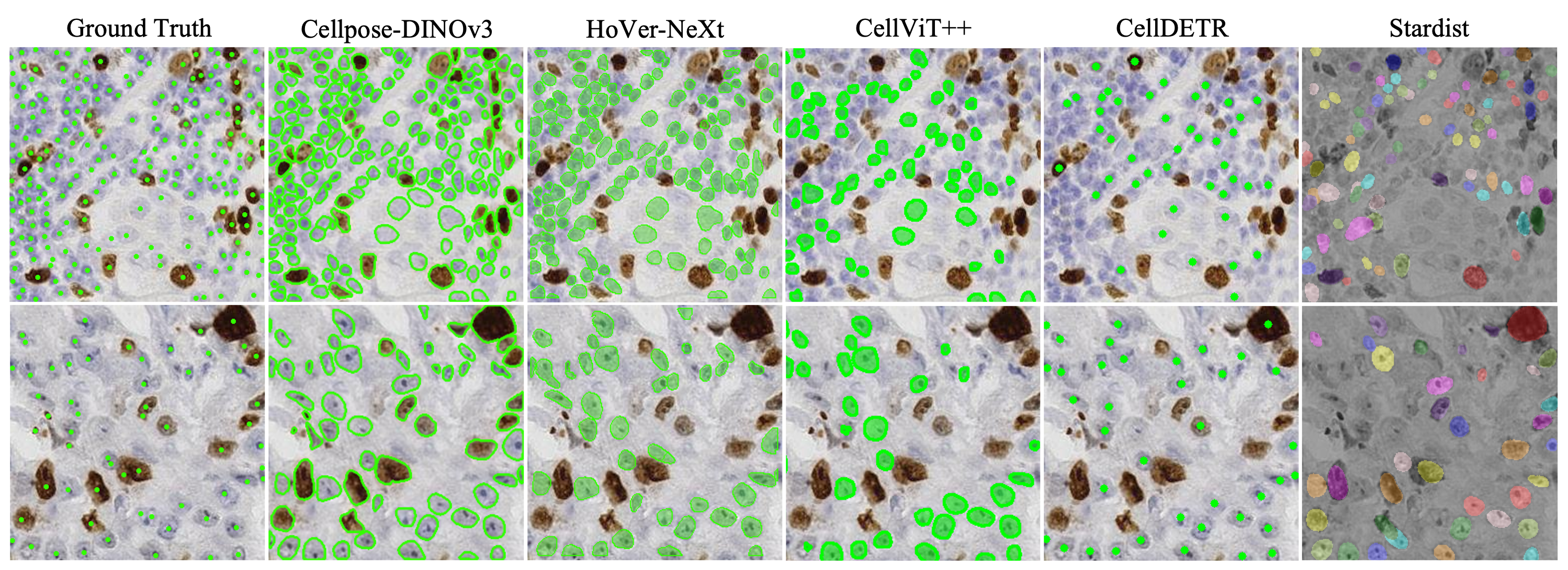}
    \caption{Qualitative comparison of nuclei segmentation and detection results on MIST dataset tiles. Results from Cellpose-DINOv3, HoVer-NeXt, CellViT++, CellDETR, and StarDist are compared against the ground-truth annotations.}
    \label{fig:segmodels_methods}
\end{figure}

Cellpose-DINOv3 achieved the most reliable nuclei detection on MIST, with the highest precision and recall and the lowest MAE. HoVer-NeXt and CellViT++, trained mainly on H\&E images where nuclei appear blue, detected hematoxylin-dominant nuclei well but frequently missed DAB-stained brown nuclei, limiting their generalization to IHC staining variability (Figure~\ref{fig:segmodels_methods}). StarDist, which relies on nuclear shape rather than color, achieved higher recall but still missed many nuclei overall. CellDETR obtained the lowest recall, indicating poor transfer to the IHC images. Given its high precision, high recall, and low counting error, Cellpose-DINOv3 was used as the detector in all subsequent experiments.

\begin{table}[ht]
    \caption{Quantitative comparison of nuclei segmentation/detection models on the MIST dataset for cell detection (all nuclei).}
    \label{tab:tab1}
    \centering
    \small
    \begin{tabular}{lccccc} 
        \toprule
        \textbf{Model} & \textbf{Precision(\%)$\uparrow$} & \textbf{Recall(\%)$\uparrow$} & \textbf{F1-Score(\%)$\uparrow$} & \textbf{MAE$\downarrow$} \\
        \midrule
        Cellpose-DINOv3 \cite{Pachitariu2025} & \textbf{91.49} & \textbf{80.26} & \textbf{85.51} & \textbf{13.82} \\ 
        HoVer-NeXt \cite{baumann2024hovernext} & 91.00 & 54.00 & 67.78 & 42.00 \\
        CellViT++ \cite{cellvit} & 91.22 & 50.11 & 64.69 & 46.84 \\
        StarDist \cite{weigert2022} & 90.27 & 63.65 & 74.66 & 33.46 \\
        CellDETR \cite{pmlr-v250-pina24a} & 90.18 & 35.57 & 51.02 & 62.55 \\
        \bottomrule
    \end{tabular}
\end{table}

\subsection{K-CLUE}
We evaluate the unsupervised Ki67-positive and Ki67-negative cell typing framework directly on RGB IHC tiles. Nuclei are detected with Cellpose-DINOv3 (Section~\ref{sec:seg_models}) using its default inference parameters. Each detected nucleus is cropped and resized to the $224\times224$ input expected by the pretrained encoders, padded with the median dataset background where the nucleus is smaller than the input window. The feature encoder is treated as an interchangeable component of the framework: all encoders are used frozen, with no fine-tuning, stain calibration, or resolution-specific adjustment, and every configuration reported below differs only in which encoder produces the nuclear embeddings.

The resulting embeddings are reduced with PCA, retaining the components that explain 90\% of the variance, and whitened (or not, depending on the experiments) as described in Section~\ref{sec:whitening}. Nuclei are then split into two clusters with $k$-means. The cluster with the higher mean DAB optical density is labeled Ki67-positive and the other Ki67-negative. This assignment rule uses no annotations, so the pipeline remains fully unsupervised end-to-end.

Since $k$-means depends on its initialization, we run it with five random seeds and 20 initializations per seed, retaining the minimum-inertia solution for each seed. All runs converge to the same cluster assignments, so we report a single result for each configuration.

Predicted nuclei are matched to ground-truth annotations using one-to-one Hungarian assignment with a $3\,\mu\mathrm{m}$ tolerance (Section~\ref{sec:containment}). For the representation-level experiments, precision, recall, and F1-score are computed separately for each class over matched nuclei only. We additionally report the Matthews correlation coefficient (MCC), computed from the full binary confusion matrix and reported as a single value per configuration. MCC ranges from $-1$ to $+1$ and is less affected by the class imbalance in MIST (1647 Ki67-positive and 4138 Ki67-negative nuclei) than the per-class F1-score.

Counting performance is evaluated independently of matching using the predicted and ground-truth cell counts in each tile. We report the mean absolute error (MAE) separately for Ki67-positive and Ki67-negative nuclei. Since negative nuclei are more numerous per tile, similar relative counting errors can result in larger absolute MAE for the negative class.

\paragraph{Handcrafted color representations.}
To assess whether explicit chromatic information is sufficient for annotation-free Ki67 typing, we additionally replace the pretrained embedding with handcrafted color descriptors while keeping the remaining pipeline unchanged. For each segmented nucleus, pixel values within the instance mask are represented in HED, HSV, and CIE L$^*$a$^*$b$^*$ color spaces. Each channel is summarized using distributional statistics over the nuclear pixels, and we also consider the concatenation of the three representations. A simple stain-based baseline represents each nucleus by its mean DAB optical density and separates the two classes with a global Otsu threshold. Handcrafted descriptors are standardized and reduced with PCA before clustering with the same global \(k\)-means procedure used for the learned representations. Appendix~\ref{A.4} compares the individual handcrafted representations with and without whitening.

\subsection{Representation geometry and whitening}
\label{sec:whitening}
This section examines how the choice of feature encoder affects annotation-free Ki67 typing, and why.  Within each convolutional family we evaluated several variants, namely ResNet-34/50/101, EfficientNet-B0/B1/B2/B3 and ConvNeXt-V2 Tiny/Base/Large. These generic architectures are compared against a self-supervised vision model, DINOv3 ViT-S/16~\cite{simeoni2025dinov3}, and against three pathology-specific encoders, HoVer-NeXt, UNI2-h~\cite{chen2024uni} and Virchow2~\cite{zimmermann2024virchow2scalingselfsupervisedmixed}. Tables \ref{tab:tab2} and \ref{tab:tab3} report one representative per convolutional family, selected on unwhitened performance. Only the encoder differs between configurations. Segmentation, PCA, cluster assignment and evaluation are identical throughout.

\begin{table}[ht]
    \caption{Comparison of pretrained encoders for Ki67 cell typing on MIST \textbf{before} PCA whitening. Classification metrics are computed over Cellpose-DINOv3 detections matched to ground-truth nuclei, while MAE reports per-tile counting error over the predicted cell set.}
    \label{tab:tab2}
    \centering
    \footnotesize
    \begin{tabular}{lcccccc} 
        \toprule
        \textbf{Model} & \textbf{Class} & \textbf{Precision(\%)$\uparrow$} & \textbf{Recall(\%)$\uparrow$} & \textbf{F1-Score(\%)$\uparrow$} & \textbf{MAE$\downarrow$} & \textbf{MCC$\uparrow$} \\
        \midrule
        \multirow{2}{*}{EfficientNet-B2} & Ki67+ & 94.06 & 94.99 & 94.52 & 5.61 & \multirow{2}{*}{0.9210} \\
        & Ki67- & 97.79 & 97.37 & 97.58 & 12.14 & \\
        \midrule
        
        \multirow{2}{*}{ResNet-101} & Ki67+ & 83.15 & 93.01 & 87.81 & 6.68 & \multirow{2}{*}{0.8229} \\
        & Ki67- & 96.76 & 91.72 & 94.18 & 14.46 & \\
        \midrule
        
        \multirow{2}{*}{ConvNeXt V2 Tiny} & Ki67+ & 94.16 & 91.11 & 92.61 & 6.41 & \multirow{2}{*}{0.8947} \\
        & Ki67- & 96.15 & 97.52 & 96.83 & 11.45 & \\
        \midrule
        
        \multirow{2}{*}{DINOv3 ViT-S/16} & Ki67+ & 90.98 & 94.71 & 92.81 & 6.36 & \multirow{2}{*}{0.8959} \\
        & Ki67- & 97.63 & 95.88 & 96.75 & 13.11 & \\
        \midrule
        
        \multirow{2}{*}{HoVer-NeXt} & Ki67+ & 39.38 & 69.87 & 50.37 & 28.38 & \multirow{2}{*}{0.2091} \\
        & Ki67- & 79.94 & 52.76 & 63.57 & 36.41 & \\
        \midrule
        
       \multirow{2}{*}{UNI2} & Ki67+ & 57.84 & 79.11 & 66.83 & 16.73 & \multirow{2}{*}{0.5023} \\
       & Ki67- & 89.06 & 74.67 & 81.23 & 22.48 & \\
       \midrule
        
       \multirow{2}{*}{Virchow2} & Ki67+ & 46.98 & 88.78 & 61.44 & 27.89 & \multirow{2}{*}{0.4172} \\
       & Ki67- & 91.91 & 55.98 & 69.58 & 36.32 & \\
       \bottomrule
   \end{tabular}
\end{table}

Before whitening, performance varies substantially across encoders, as Table~\ref{tab:tab2} shows. EfficientNet-B2 reaches a Ki67-positive F1-score of 94.52\% and an MCC of 0.921, whereas HoVer-NeXt performs close to chance at 50.37\% and 0.209, a spread of 44.15 F1-score points and 0.712 in MCC. The remaining encoders are distributed across this range, with Virchow2 at 61.44\%, UNI2 at 66.83\% and ResNet-101 at 87.81\%. This ranking follows neither encoder capacity nor pretraining domain, since the three pathology-specific encoders occupy the bottom three positions despite being trained on considerably more histopathology data. The failure modes also differ in kind. Virchow2 assigns too many nuclei to the positive cluster, with a precision of 46.98\% at a recall of 88.78\%, while HoVer-NeXt assigns nearly half the negative population to the positive cluster, with a Ki67-negative recall of 52.76\%. In both cases the recovered clusters correspond to something other than Ki67 status.

As described in Section~\ref{sec:embeddings}, components with larger variance dominate the Euclidean distance, so the clusters follow the attribute those components encode. To remove this dependence we apply PCA whitening before clustering. Table~\ref{tab:tab3} reports the results.

\begin{table}[ht]
    \caption{Comparison of pretrained encoders for Ki67 cell typing on MIST \textbf{after} PCA whitening. Classification metrics are computed over Cellpose-DINOv3 detections matched to ground-truth nuclei, while MAE reports per-tile counting error over the predicted cell set.}
    \label{tab:tab3}
    \centering
    \footnotesize
    \begin{tabular}{lcccccc} 
        \toprule
        \textbf{Model} & \textbf{Class} & \textbf{Precision(\%)$\uparrow$} & \textbf{Recall(\%)$\uparrow$} & \textbf{F1-Score(\%)$\uparrow$} & \textbf{MAE$\downarrow$} & \textbf{MCC$\uparrow$} \\
        \midrule
        \multirow{2}{*}{EfficientNet-B2} & Ki67+ & 96.01 & 93.37 & 94.67 & 5.75 & \multirow{2}{*}{0.9239} \\
        & Ki67- & 97.12 & 98.30 & 97.70 & 11.36 & \\
        \midrule
        
        \multirow{2}{*}{ResNet-101} & Ki67+ & 94.80 & 94.00 & 94.40 & 5.62 & \multirow{2}{*}{0.9196} \\
        & Ki67- & 97.37 & 97.74 & 97.56 & 11.91 & \\
        \midrule
        
        \multirow{2}{*}{ConvNeXt V2 Tiny} & Ki67+ & 95.42 & 94.14 & 94.78 & 5.52 & \multirow{2}{*}{0.9251} \\
        & Ki67- & 97.44 & 98.02 & 97.73 & 11.77 & \\
        \midrule
        
        \multirow{2}{*}{DINOv3 ViT-S/16} & Ki67+ & 95.67 & 93.51 & 94.58 & 5.66 & \multirow{2}{*}{0.9224} \\
        & Ki67- & 97.18 & 98.14 & 97.66 & 11.59 & \\
        \midrule
        
        \multirow{2}{*}{HoVer-NeXt} & Ki67+ & 94.02 & 94.35 & 94.19 & 5.18 & \multirow{2}{*}{0.9163} \\
        & Ki67- & 97.52 & 97.37 & 97.44 & 12.43 & \\
        \midrule
        
       \multirow{2}{*}{UNI2} & Ki67+ & 96.12 & 92.59 & 94.32 & 6.00 & \multirow{2}{*}{0.9193} \\
       & Ki67- & 96.80 & 98.36 & 97.57 & 11.14 & \\
       \midrule
        
       \multirow{2}{*}{Virchow2} & Ki67+ & 95.27 & 92.38 & 93.80 & 5.88 & \multirow{2}{*}{0.9116} \\
       & Ki67- & 96.70 & 97.99 & 97.34 & 11.52 & \\
       \bottomrule
   \end{tabular}
\end{table}

Whitening reduces the differences between encoders, and the largest gains occur where performance was worst. The Ki67-positive F1-score rises from 50.37\% to 94.19\% for HoVer-NeXt, from 61.44\% to 93.80\% for Virchow2, from 66.83\% to 94.32\% for UNI2 and from 87.81\% to 94.40\% for ResNet-101, while the counting error of HoVer-NeXt falls from an MAE of 28.38 to 5.18. The encoders that already clustered well change only marginally, so whitening is close to neutral where the stain already occupies the leading directions. A spread of 44.15 points becomes a spread of 1.00 points, and the seven encoders differ by 0.014 in MCC. Annotation-free Ki67 typing is therefore not a property of any particular pretrained encoder. What distinguishes them is whether the Ki67 signal occupies the directions along which $k$-means separates the data. PCA whitening is used in all subsequent experiments.

The low unwhitened scores could instead indicate that some embeddings do not reflect Ki67 status on its main components. To test this, we apply the linear probe of Section~\ref{sec:embeddings} to each frozen representation. Every encoder reaches an AUC above 0.98 in Table~\ref{tab:tab4}, including HoVer-NeXt at 0.9925, so Ki67 status is linearly decodable from all seven representations and the differences in Table~\ref{tab:tab2} cannot be explained by absent information. Decodability does not predict clustering either, since the Spearman correlation between the probe AUC and the unwhitened Ki67-positive F1-score is $-0.54$. The two highest probe AUCs, Virchow2 at 0.9964 and UNI2 at 0.9957, belong to two of the three worst unwhitened performers. Dimensionality is equally uninformative, as DINOv3 ViT-S/16 uses 384 dimensions and matches encoders using up to 2560. The whitened results also approach this supervised ceiling, standing on average 0.9 points below the probe trained on the same embeddings and above it for one encoder.

\begin{table}[ht]
    \caption{Supervised linear probe performance on the frozen embeddings of each encoder, using ground-truth labels of the 4643 matched nuclei with cross-validation grouped by tile ID. AUC and AP denote the area under the ROC curve and the average precision. F1-Score + is the Ki67-positive F1-score.}
    \label{tab:tab4}
    \centering
    \small
    \begin{tabular}{lccccc}
        \toprule
        \textbf{Model} & \textbf{Dim.} & \textbf{AUC$\uparrow$} & \textbf{AP$\uparrow$} & \textbf{F1-Score + (\%)$\uparrow$} & \textbf{Bal. acc.(\%)$\uparrow$} \\
        \midrule
        EfficientNet-B2   & 1408 & 0.9902 & 0.9813 & 95.22 & 96.79 \\
        ResNet-101        & 2048 & 0.9833 & 0.9641 & 93.01 & 95.77 \\
        ConvNeXt-V2 Tiny  & 768  & 0.9926 & 0.9869 & 95.53 & 97.21 \\
        DINOv3 ViT-S/16   & 384  & 0.9911 & 0.9757 & 95.51 & 97.09 \\
        HoVer-NeXt        & 1024 & 0.9925 & 0.9852 & 95.77 & 97.34 \\
        UNI2              & 1536 & 0.9957 & 0.9924 & 95.94 & 97.45 \\
        Virchow2          & 2560 & 0.9964 & 0.9925 & 95.87 & 97.42 \\
        \bottomrule
    \end{tabular}
\end{table}

To identify what the leading directions encode, we correlated the top principal components with per-nucleus covariates measured directly from the image, namely the mean DAB and hematoxylin optical densities, the mean saturation and luminance, the nuclear area and the tile of origin. No Ki67 annotations are used in this analysis. Table~\ref{tab:tab5} reports the association of the first principal component with each covariate.

\begin{table}[ht]
    \caption{Absolute Pearson correlation between the first principal component of each frozen representation and per-nucleus covariates measured from the image. The tile column reports the correlation ratio $\eta$, as tile of origin is categorical.}
    \label{tab:tab5}
    \centering
    \small
    \begin{tabular}{lcccccc}
        \toprule
        \textbf{Model} & \textbf{DAB} & \textbf{Hematoxylin} & \textbf{Saturation} & \textbf{Luminance} & \textbf{Area} & \textbf{Tile} \\
        \midrule
        EfficientNet-B2 & 0.829 & 0.118 & 0.747 & 0.774 & 0.488 & 0.626 \\
        ResNet-101 & 0.728 & 0.271 & 0.584 & 0.595 & 0.639 & 0.586 \\
        ConvNeXt-V2 Tiny & 0.854 & 0.162 & 0.753 & 0.758 & 0.532 & 0.594 \\
        DINOv3 ViT-S/16 & 0.774 & 0.092 & 0.683 & 0.732 & 0.568 & 0.593 \\
        HoVer-NeXt & 0.238 & 0.121 & 0.335 & 0.234 & 0.654 & 0.501 \\
        UNI2 & 0.579 & 0.105 & 0.574 & 0.628 & 0.422 & 0.634 \\
        Virchow2 & 0.550 & 0.432 & 0.652 & 0.763 & 0.138 & 0.736 \\
        \bottomrule
    \end{tabular}
\end{table}

For EfficientNet-B2, ConvNeXt-V2 Tiny, DINOv3 ViT-S/16 and ResNet-101, PC1 is most strongly associated with DAB optical density, with values between 0.728 and 0.854. For the three remaining encoders it is not. The PC1 of HoVer-NeXt is dominated by nuclear area, consistent with its Ki67-negative recall of 52.76\%, that of Virchow2 by luminance together with the highest hematoxylin association of any encoder, which is consistent with a precision of 46.98\% at a recall of 88.78\%, and that of UNI2 tile identity rather than per-nucleus staining. %The same separation holds when luminance is held fixed, with partial correlations between 0.426 and 0.610 for the four encoders whose PC1 is dominated by DAB and between 0.076 and 0.280 for the other three. 
Across the seven encoders, the association between PC1 and DAB predicts the unwhitened Ki67-positive F1-score with a Pearson correlation of 0.955. The clustering therefore recovers the structure that occupies the leading direction of the representation, and not Ki67 status. Appendix~\ref{A.5} extends the analysis to the first ten components and shows that the stain signal is displaced rather than absent, since the second component of HoVer-NeXt is associated with DAB optical density at 0.838, against 0.238 for the first.

To examine how whitening produces this effect, we divided each retained component by its standard deviation raised to the power $\alpha$, so that $\alpha = 0$ leaves the variance structure unchanged and $\alpha = 1$ gives every component unit variance. Appendix~\ref{A.6} reports the Ki67-positive F1-score across this range. The encoders that fail without whitening remain at their initial values over most of the range and then recover more than 30 points in a single step. Whitening therefore adds no information to the representation, since partial suppression of the dominant nuisance direction leaves the clustering unchanged until that direction no longer dominates the Euclidean distance, at which point the correct clusters appear.

\subsection{Results}
\label{sec:results}
\begin{figure}[ht]
    \centering
    \includegraphics[width=1.0\linewidth]{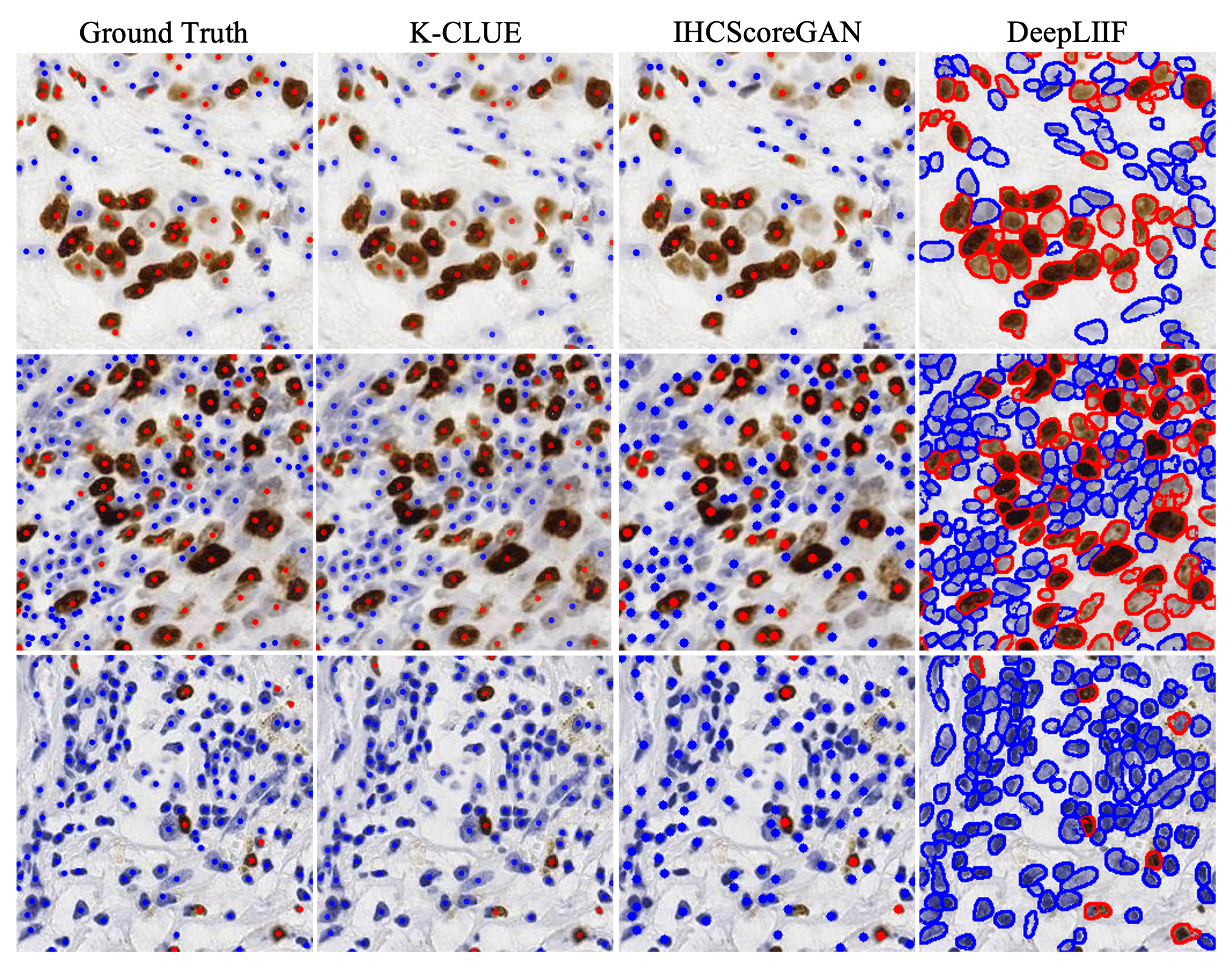}
    \caption{Qualitative comparison of K-CLUE, IHCScoreGAN, and DeepLIIF for Ki67-positive (red) and Ki67-negative (blue) cell detection on representative MIST tiles.}
    \label{fig:mist_results}
\end{figure}

We compare K-CLUE with DeepLIIF, IHCScoreGAN, and two handcrafted color baselines (Table~\ref{tab:results1} and Figure~\ref{fig:mist_results}). The DAB baseline assigns each segmented nucleus its mean DAB optical density and separates Ki67-positive and Ki67-negative nuclei using a global Otsu threshold. HED+HSV+L$^*$a$^*$b$^*$ combines complementary stain and color information, and we use it for both datasets since concatenating the three color spaces avoids selecting one per cohort and matches the single-configuration setting used for K-CLUE. Its features are standardized and PCA-whitened (Appendix~\ref{A.4}). DeepLIIF is trained from paired IHC and mpIF images, while IHCScoreGAN is designed for IHC images at a fixed spatial resolution. K-CLUE requires neither paired data nor task-specific supervision.

\begin{table}[ht]
    \centering
    \footnotesize
    \caption{End-to-end Ki67-positive and Ki67-negative cell detection and classification on MIST using Hungarian matching with a $3\,\mu\mathrm{m}$ tolerance.}
    \label{tab:results1}
    \begin{tabular}{l cccc cccc}
        \toprule
        & \multicolumn{4}{c}{Ki67-positive cells} & \multicolumn{4}{c}{Ki67-negative cells} \\ 
        \cmidrule(lr){2-5} \cmidrule(lr){6-9}
        \textbf{Model} & \textbf{P(\%)$\uparrow$} & \textbf{R(\%)$\uparrow$} & \textbf{F1(\%)$\uparrow$} & \textbf{MAE$\downarrow$} &
        \textbf{P(\%)$\uparrow$} & \textbf{R(\%)$\uparrow$} & \textbf{F1(\%)$\uparrow$} & \textbf{MAE$\downarrow$} \\
        \midrule
        DAB baseline & \textbf{96.86} & 61.75 & 75.42 & 10.84 & 80.15 & 77.96 & 79.04 & \textbf{9.34} \\

        HED+HSV+L$^*$a$^*$b$^*$ & 91.19 & 77.90 & 84.02 & 6.36 & 86.72 & 76.87 & 81.50 & 10.64 \\
        
        DeepLIIF & 71.40 & 77.78 & 74.46 & 6.09 & 61.50 & \textbf{78.42} & 68.94 & 25.61 \\ 
        
        IHCScoreGAN & 84.88 & 63.75 & 72.82 & 7.82 & 66.43 & 70.86 & 68.57 & 21.07 \\ 
        
        K-CLUE & 89.41 & \textbf{81.00} & \textbf{85.00} & \textbf{5.52} & \textbf{88.25} & 76.41 & \textbf{81.91} & 11.77 \\
        \bottomrule
    \end{tabular}
\end{table}

Following the analysis of Section~\ref{sec:whitening}, we use ConvNeXt-V2 Tiny as the default encoder for the end-to-end comparisons. After whitening, performance differs by at most one F1 point across the seven evaluated encoders. K-CLUE achieves the highest end-to-end F1-score for both Ki67-positive and Ki67-negative nuclei, with 85.00\% and 81.91\%, respectively. It also attains the highest recall for Ki67-positive cells (81.00\%) and the highest precision for Ki67-negative cells (88.25\%). The combined handcrafted baseline is competitive, reaching F1-scores of 84.02\% and 81.50\%, while the DAB baseline shows high precision but substantially lower recall for Ki67-positive nuclei. 

K-CLUE also obtains the lowest MAE for Ki67-positive cells, while the DAB baseline gives the lowest Ki67-negative MAE. Since the evaluation is end-to-end, missed nuclei and unmatched predictions contribute directly to false negatives and false positives, so the final performance also reflects the quality of the initial segmentation.

The strong performance of the handcrafted representation confirms that Ki67 status is largely reflected in chromatic information. However, these features explicitly encode stain and color statistics chosen for this task. K-CLUE recovers comparable separation from pretrained representations without a task-specific color space or handcrafted descriptor.

\subsection{Evaluation on an external dataset}
\label{sec:generalization}
We evaluate the methods on BCData, an external dataset acquired under different imaging conditions and at a higher spatial resolution than MIST (Table~\ref{tab:results2}), using the same unsupervised procedure. End-to-end cell detection and classification are evaluated with the same protocol as on MIST, using Hungarian matching with a $3\,\mu\mathrm{m}$ tolerance.

As on MIST, the crop size is set to the largest detected nucleus plus one pixel of padding, resulting in $126\times126$ crops at the resolution of BCData. An additional ablation comparing all pretrained encoders on BCData is reported in Appendix~\ref{A.7}, using the same matched-nucleus evaluation protocol as on MIST.

\begin{table}[ht]
    \centering
    \footnotesize
    \caption{End-to-end Ki67-positive and Ki67-negative cell detection and classification on BCData using Hungarian matching with a $3\,\mu\mathrm{m}$ tolerance.}
    \label{tab:results2}
    \begin{tabular}{l cccc cccc}
        \toprule
        & \multicolumn{4}{c}{Ki67-positive cells} & \multicolumn{4}{c}{Ki67-negative cells} \\ 
        \cmidrule(lr){2-5} \cmidrule(lr){6-9}
        \textbf{Model} & \textbf{P(\%)$\uparrow$} & \textbf{R(\%)$\uparrow$} & \textbf{F1(\%)$\uparrow$} & \textbf{MAE$\downarrow$} &
        \textbf{P(\%)$\uparrow$} & \textbf{R(\%)$\uparrow$} & \textbf{F1(\%)$\uparrow$} & \textbf{MAE$\downarrow$} \\
        \midrule
        DAB baseline & \textbf{82.48} & 70.45 & 75.99 & 2.28 & 77.81 & 80.68 & 79.22 & 4.39 \\

        HED+HSV+L$^*$a$^*$b$^*$ & 78.76 & 81.97 & \textbf{80.33} & \textbf{2.02} & 84.08 & 79.58 & \textbf{81.77} & \textbf{4.08} \\
        
        DeepLIIF & 55.05 & \textbf{87.71} & 67.65 & 5.86 & 64.86 & \textbf{83.60} & 73.05 & 6.73 \\ 

        IHCScoreGAN & 72.92 & 77.13 & 74.96 & 2.22 & 77.17 & 67.00 & 71.73 & 4.51 \\ 
        
        K-CLUE & 73.94 & 86.17 & 79.59 & 2.69 & \textbf{86.38} & 76.53 & 81.16 & 4.45 \\
        \bottomrule
    \end{tabular}
\end{table}

K-CLUE achieves an F1-score of 79.59\% for Ki67-positive nuclei, slightly below the combined handcrafted baseline at 80.33\%, and substantially above DeepLIIF and IHCScoreGAN. For Ki67-negative nuclei, the combined HED+HSV+L$^*$a$^*$b$^*$ representation obtains the highest F1-score in Table~\ref{tab:results2} at 81.77\%, followed closely by K-CLUE at 81.16\%. K-CLUE reaches the highest precision for Ki67-negative cells (86.38\%), while DeepLIIF attains the highest recall for both classes but with considerably lower precision. The handcrafted representation also gives the lowest counting error for both classes, with MAE of 2.02 and 4.08. 

Overall, without target labels or supervised adaptation, K-CLUE stays within one F1 point of the handcrafted baseline. A paired bootstrap over tiles (10{,}000 resamples) shows that the 95\% confidence intervals of the F1 differences (K-CLUE minus handcrafted baseline) include zero on MIST ($[-0.04, 2.11]$ and $[-0.04, 0.79]$ percentage points for Ki67-positive and Ki67-negative nuclei), so the two methods are not significantly different. On BCData, K-CLUE performs slightly below the handcrafted baseline, with differences of $-0.74$ and $-0.61$ percentage points (95\% CIs: $[-1.45, -0.05]$ and $[-1.02, -0.23]$).

Among the individual color spaces (Appendix~\ref{A.4}), L$^*$a$^*$b$^*$ is marginally above K-CLUE on BCData (80.25\% and 81.96\%), while HSV, the best for Ki67-negative cells on MIST, drops to 57.65\% and 68.09\%. The best color space thus differs between datasets, a selection that both the combined descriptor and K-CLUE avoid.

\section{Conclusion}
%We presented K-CLUE, an annotation-free framework for Ki67-positive and Ki67-negative cell typing from IHC images. The method combines nuclei segmentation, frozen pretrained representations, PCA whitening, and unsupervised clustering, without task-specific training or target-domain annotations.

We presented K-CLUE, an annotation-free framework for Ki67-positive and Ki67-negative cell typing from IHC images that combines nuclei segmentation, frozen pretrained representations, PCA whitening, and unsupervised clustering, without task-specific training or target-domain annotations.

Our experiments show that the choice of pretrained representation has a large effect on naive clustering, despite Ki67 status being linearly decodable from all evaluated encoders. PCA whitening substantially reduces this dependence by normalizing differences in the variance structure of the embedding spaces, bringing the performance of the seven encoders to within one F1-score point on MIST. Whitened handcrafted color representations perform comparably to K-CLUE, confirming that Ki67 separation is strongly driven by chromatic information, but unlike K-CLUE they require a task-specific color space and descriptor.

In end-to-end evaluation, K-CLUE remains competitive with DeepLIIF and IHCScoreGAN on both MIST and BCData without supervised adaptation to either dataset. Its main limitation is the dependence on the upstream nuclei segmentation, since missed and spurious detections directly affect the final cell-typing performance. Future work will extend the framework to distinguish tumor cells, lymphocytes, and other cellular populations, allowing Ki67 proliferation indices to be estimated within clinically relevant tissue compartments.

\subsection*{Acknowledgments}
This work was supported by national funds through Fundação para a Ciência e a Tecnologia, I.P. (FCT) under projects UID/50021/2025 (DOI: https://doi.org/10.54499/UID/50021/2025) and UID/PRR/50021/2025 (DOI: https://doi.org/10.54499/UID/PRR/50021/2025), and Center for Responsible AI (C628696807-00454142).

\bibliography{refs}

@article{Santos2025,
  author = {Santos, Talita Alves do Nascimento and Silva, Anna Karoline Fausto da and Cunha, Karin Soares Gonçalves and Costa, Camila Braz Pereira da and Silva, Aldo Rodrigues da and Castro, Helena Carla and Oliveira, Nathália Silva Carlos},
  title = {{Current advances in cancer immunohistochemistry: a new perspective for the Ki-67 biomarker}},
  journal = {ecancermedicalscience},
  year = {2025},
  volume = {19},
  pages = {1863},
  month = {03},
  doi = {10.3332/ecancer.2025.1863},
}

@article{Davey2021,
  author = {Davey, Matthew G. and Hynes, Jarlath P. and Kerin, Michael J. and Miller, Nicola and Lowery, Aoife J.},
  title = {{Ki-67 as a Prognostic Biomarker in Invasive Breast Cancer}},
  journal = {Cancers},
  year = {2021},
  volume = {13},
  number = {17},
  pages = {4455},
  month = {09},
  doi = {10.3390/cancers13174455},
}

@article{Aung2021,
  author = {Aung, Thazin Nwe and Acs, Balazs and Warrell, Jonathan and Bai, Yalai and Gaule, Patricia and Martinez-Morilla, Sandra and Vathiotis, Ioannis and Shafi, Saba and Moutafi, Myrto and Gerstein, Mark and Freiberg, Benjamin and Fulton, Regan and Rimm, David L.},
  title = {{A new tool for technical standardization of the Ki67 immunohistochemical assay}},
  journal = {Modern Pathology},
  year = {2021},
  volume = {34},
  number = {7},
  pages = {1261--1270},
  month = {07},
  doi = {10.1038/s41379-021-00745-6},
}

@article{Dy2024,
  author = {Dy, Amanda and Nguyen, Ngoc-Nhu Jennifer and Meyer, Julien and Dawe, Melanie and Shi, Wei and Androutsos, Dimitri and Fyles, Anthony and Liu, Fei-Fei and Done, Susan and Khademi, April},
  title = {{AI improves accuracy, agreement and efficiency of pathologists for Ki67 assessments in breast cancer}},
  journal = {Scientific Reports},
  year = {2024},
  volume = {14},
  number = {1},
  pages = {1283},
  month = {01},
  doi = {10.1038/s41598-024-51723-2},
}

@article{AlTarawneh2025,
  author = {Al-Tarawneh, Zakaria A. and Tarawneh, Ahmad S. and Mbaidin, Almoutaz and Fern{\'a}ndez-Delgado, Manuel and G{\'a}ndara-Vila, Pilar and Hassanat, Ahmad and Cernadas, Eva},
  title = {Cell Detection in Biomedical Immunohistochemical Images Using Unsupervised Segmentation and Deep Learning},
  journal = {Electronics},
  year = {2025},
  volume = {14},
  number = {18},
  pages = {3705},
  doi = {10.3390/electronics14183705},
}

@InProceedings{ihcscoregan,
  title = {{IHCScoreGAN: An unsupervised generative adversarial network for end-to-end ki67 scoring for clinical breast cancer diagnosis}},
  author = {Molnar, Carl and Tavolara, Thomas E. and Garcia, Christopher A. and McClintock, David S. and Zarella, Mark D. and Han, Wenchao},
  booktitle = {Proceedings of The 7th International Conference on Medical Imaging with Deep Learning},
  pages = {1011--1025},
  year = {2024},
  editor = {Burgos, Ninon and Petitjean, Caroline and Vakalopoulou, Maria and Christodoulidis, Stergios and Coupe, Pierrick and Delingette, Hervé and Lartizien, Carole and Mateus, Diana},
  volume = {250},
  series = {Proceedings of Machine Learning Research},
  month = {03--05 Jul},
  publisher = {PMLR},
}

@article{Pachitariu2025,
  author = {Pachitariu, Marius and Rariden, Michael and Stringer, Carsen},
  title = {{Cellpose-SAM: superhuman generalization for cellular segmentation}},
  journal = {bioRxiv},
  year = {2025},
  doi = {10.1101/2025.04.28.651001},
  publisher = {Cold Spring Harbor Laboratory}
}

@article{Ghahremani2022,
  author = {Ghahremani, Parmida and Li, Yanyun and Kaufman, Arie and Vanguri, Rami and Greenwald, Noah and Angelo, Michael and Hollmann, Travis J. and Nadeem, Saad},
  title = {{Deep Learning-Inferred Multiplex ImmunoFluorescence for Immunohistochemical Image Quantification}},
  journal = {Nature Machine Intelligence},
  year = {2022},
  volume = {4},
  number = {4},
  pages = {401--412},
  doi = {10.1038/s42256-022-00471-x},
  note = {PMID: 36118303}
}

@misc{cellvit,
  title={{CellViT++: Energy-Efficient and Adaptive Cell Segmentation and Classification Using Foundation Models}}, 
  author={Fabian Hörst and Moritz Rempe and Helmut Becker and Lukas Heine and Julius Keyl and Jens Kleesiek},
  year={2025},
  eprint={2501.05269},
  archivePrefix={arXiv},
  primaryClass={cs.CV},
}

@inproceedings{baumann2024hovernext,
  title={{HoVer-NeXt: A Fast Nuclei Segmentation and Classification Pipeline for Next Generation Histopathology}},
  author={Elias Baumann and Bastian Dislich and Josef Lorenz Rumberger and Iris D. Nagtegaal and Maria Rodriguez Martinez and Inti Zlobec},
  booktitle={Medical Imaging with Deep Learning},
  year={2024},
}

@inproceedings{weigert2022,
  author = {Martin Weigert and Uwe Schmidt},
  title = {Nuclei Instance Segmentation and Classification in Histopathology Images with StarDist},
  booktitle = {The IEEE International Symposium on Biomedical Imaging Challenges (ISBIC)},
  year = {2022},
  doi = {10.1109/ISBIC56247.2022.9854534}
}

@InProceedings{pmlr-v250-pina24a,
  title = {{Cell-DETR: Efficient cell detection and classification in WSIs with transformers}},
  author = {Pina, Oscar and Dorca, Eduard and Vilaplana, Veronica},
  booktitle = {Proceedings of The 7th International Conference on Medical Imaging with Deep Learning},
  pages = {1128--1141},
  year = {2024},
  editor = {Burgos, Ninon and Petitjean, Caroline and Vakalopoulou, Maria and Christodoulidis, Stergios and Coupe, Pierrick and Delingette, Hervé and Lartizien, Carole and Mateus, Diana},
  volume = {250},
  series = {Proceedings of Machine Learning Research},
  month = {03--05 Jul},
  publisher = {PMLR},
}

@ARTICLE{10.3389/fbioe.2019.00226, 
  AUTHOR={Geread, Rokshana S.  and Morreale, Peter  and Dony, Robert D.  and Brouwer, Emily  and Wood, Geoffrey A.  and Androutsos, Dimitrios  and Khademi, April },  
  TITLE={{IHC Color Histograms for Unsupervised Ki67 Proliferation Index Calculation}},    
  JOURNAL={Frontiers in Bioengineering and Biotechnology},     
  VOLUME={Volume 7 - 2019},
  YEAR={2019},
  DOI={10.3389/fbioe.2019.00226},
  ISSN={2296-4185},
}

@InProceedings{Kirillov_2023_ICCV,
    author    = {Kirillov, Alexander and Mintun, Eric and Ravi, Nikhila and Mao, Hanzi and Rolland, Chloe and Gustafson, Laura and Xiao, Tete and Whitehead, Spencer and Berg, Alexander C. and Lo, Wan-Yen and Dollar, Piotr and Girshick, Ross},
    title     = {Segment Anything},
    booktitle = {Proceedings of the IEEE/CVF International Conference on Computer Vision (ICCV)},
    month     = {October},
    year      = {2023},
    pages     = {4015-4026}
}

@inproceedings{li2023adaptive,
  title={{Adaptive supervised patchnce loss for learning H\&E-to-IHC stain translation with inconsistent groundtruth image pairs}},
  author={Li, Fangda and Hu, Zhiqiang and Chen, Wen and Kak, Avinash},
  booktitle={International Conference on Medical Image Computing and Computer-Assisted Intervention},
  pages={632--641},
  year={2023},
  organization={Springer}
}

@InProceedings{10.1007/978-3-030-59722-1_28,
  author={Huang, Zhongyi and Ding, Yao and Song, Guoli and Wang, Lin and Geng, Ruizhe and He, Hongliang and Du, Shan and Liu, Xia and Tian, Yonghong and Liang, Yongsheng and Zhou, S. Kevin and Chen, Jie},
  editor={Martel, Anne L. and Abolmaesumi, Purang and Stoyanov, Danail and Mateus, Diana and Zuluaga, Maria A. and Zhou, S. Kevin and Racoceanu, Daniel and Joskowicz, Leo},
  title={{BCData: A Large-Scale Dataset and Benchmark for Cell Detection and Counting}},
  booktitle={Medical Image Computing and Computer Assisted Intervention -- MICCAI 2020},
  year={2020},
  publisher={Springer International Publishing},
  address={Cham},
  pages={289--298},
}

@InProceedings{pmlr-v97-tan19a,
  title     = {EfficientNet: Rethinking Model Scaling for Convolutional Neural Networks},
  author    = {Tan, Mingxing and Le, Quoc V.},
  booktitle = {Proceedings of the 36th International Conference on Machine Learning},
  pages     = {6105--6114},
  year      = {2019},
  editor    = {Chaudhuri, Kamal and Salakhutdinov, Ruslan},
  volume    = {97},
  series    = {Proceedings of Machine Learning Research},
  month     = {09--15 Jun},
  publisher = {PMLR},
}

@article{chen2024uni,
  title={Towards a General-Purpose Foundation Model for Computational Pathology},
  author={Chen, Richard J and Ding, Tong and Lu, Ming Y and Williamson, Drew FK and Jaume, Guillaume and Chen, Bowen and Zhang, Andrew and Shao, Daniel and Song, Andrew H and Shaban, Muhammad and others},
  journal={Nature Medicine},
  publisher={Nature Publishing Group},
  year={2024}
}

@misc{zimmermann2024virchow2scalingselfsupervisedmixed,
      title={Virchow2: Scaling Self-Supervised Mixed Magnification Models in Pathology}, 
      author={Eric Zimmermann and Eugene Vorontsov and Julian Viret and Adam Casson and Michal Zelechowski and George Shaikovski and Neil Tenenholtz and James Hall and David Klimstra and Razik Yousfi and Thomas Fuchs and Nicolo Fusi and Siqi Liu and Kristen Severson},
      year={2024},
      eprint={2408.00738},
      archivePrefix={arXiv},
      primaryClass={cs.CV},
}

@misc{simeoni2025dinov3,
  title={{DINOv3}},
  author={Sim{\'e}oni, Oriane and Vo, Huy V. and Seitzer, Maximilian and Baldassarre, Federico and Oquab, Maxime and Jose, Cijo and Khalidov, Vasil and Szafraniec, Marc and Yi, Seungeun and Ramamonjisoa, Micha{\"e}l and Massa, Francisco and Haziza, Daniel and Wehrstedt, Luca and Wang, Jianyuan and Darcet, Timoth{\'e}e and Moutakanni, Th{\'e}o and Sentana, Leonel and Roberts, Claire and Vedaldi, Andrea and Tolan, Jamie and Brandt, John and Couprie, Camille and Mairal, Julien and J{\'e}gou, Herv{\'e} and Labatut, Patrick and Bojanowski, Piotr},
  year={2025},
  eprint={2508.10104},
  archivePrefix={arXiv},
  primaryClass={cs.CV},
}

@article{ruifrok2001color,
  author  = {Ruifrok, Arnout C. and Johnston, Dennis A.},
  title   = {Quantification of Histochemical Staining by Color Deconvolution},
  journal = {Analytical and Quantitative Cytology and Histology},
  volume  = {23},
  number  = {4},
  pages   = {291--299},
  year    = {2001},
  month   = aug,
  pmid    = {11531144}
}

@inproceedings{he2016deep,
  title={Deep residual learning for image recognition},
  author={He, Kaiming and Zhang, Xiangyu and Ren, Shaoqing and Sun, Jian},
  booktitle={Proceedings of the IEEE conference on computer vision and pattern recognition},
  pages={770--778},
  year={2016}
}

@INPROCEEDINGS{10205236,
    author = {Woo, Sanghyun and Debnath, Shoubhik and Hu, Ronghang and Chen, Xinlei and Liu, Zhuang and Kweon, In So and Xie, Saining },
    booktitle = { 2023 IEEE/CVF Conference on Computer Vision and Pattern Recognition (CVPR) },
    title = {{ ConvNeXt V2: Co-designing and Scaling ConvNets with Masked Autoencoders }},
    year = {2023},
    volume = {},
    ISSN = {},
    pages = {16133-16142},
    doi = {10.1109/CVPR52729.2023.01548},
    publisher = {IEEE Computer Society},
    address = {Los Alamitos, CA, USA},
    month =Jun}

@InProceedings{10.1007/978-3-642-33709-3_55,
    author="J{\'e}gou, Herv{\'e}
    and Chum, Ond{\v{r}}ej",
    editor="Fitzgibbon, Andrew
    and Lazebnik, Svetlana
    and Perona, Pietro
    and Sato, Yoichi
    and Schmid, Cordelia",
    title={Negative Evidences and Co-occurences in Image Retrieval: The Benefit of PCA and Whitening},
    booktitle="Computer Vision -- ECCV 2012",
    year="2012",
    publisher="Springer Berlin Heidelberg",
    address="Berlin, Heidelberg",
    pages="774--787",
    isbn="978-3-642-33709-3"
}

@misc{su2021,
      title={Whitening Sentence Representations for Better Semantics and Faster Retrieval}, 
      author={Jianlin Su and Jiarun Cao and Weijie Liu and Yangyiwen Ou},
      year={2021},
      eprint={2103.15316},
      archivePrefix={arXiv},
      primaryClass={cs.CL},
}

@inproceedings{huang-etal-2021-whiteningbert-easy,
    title = "{W}hitening{BERT}: An Easy Unsupervised Sentence Embedding Approach",
    author = "Huang, Junjie  and
      Tang, Duyu  and
      Zhong, Wanjun  and
      Lu, Shuai  and
      Shou, Linjun  and
      Gong, Ming  and
      Jiang, Daxin  and
      Duan, Nan",
    editor = "Moens, Marie-Francine  and
      Huang, Xuanjing  and
      Specia, Lucia  and
      Yih, Scott Wen-tau",
    booktitle = "Findings of the Association for Computational Linguistics: EMNLP 2021",
    month = nov,
    year = "2021",
    address = "Punta Cana, Dominican Republic",
    publisher = "Association for Computational Linguistics",
    doi = "10.18653/v1/2021.findings-emnlp.23",
    pages = "238--244",
}

@misc{saraswatula2025,
      title={Data Whitening Improves Sparse Autoencoder Learning}, 
      author={Ashwin Saraswatula and David Klindt},
      year={2025},
      eprint={2511.13981},
      archivePrefix={arXiv},
      primaryClass={cs.LG}, 
}

\newpage

\appendix
\section{Supplementary Experimental Results}
\label{appendix}
This appendix provides supplementary experiments and analyses supporting the main text.

\subsection{Segmentation network details}
\label{A.1}
All inference parameters were kept at their default values (diameter=None, flow threshold=0.4, cell probability threshold=0.0). Table~\ref{tab:app1} compares the four Cellpose variants (Cellpose-SAM, Cellpose-SAMv2, Cellpose-DINOv3 ViT-B and Cellpose-DINOv3 ViT-L) under this setting. 

\begin{table}[ht]
    \caption{Detection performance of the four Cellpose variants on the MIST dataset using default inference parameters, against 5785 ground-truth annotations. Detected denotes the total number of predicted nuclei, TP, FP, FN, precision, recall, and F1-score are computed after matching predicted nuclei to the annotations (Hungarian matching with a $3\,\mu\mathrm{m}$ tolerance), and MAE from the per-image cell counts.}
    \label{tab:app1}
    \centering
    \footnotesize
    \begin{tabular}{lcccccccc} 
        \toprule
        \textbf{Model} & \textbf{Detected} & \textbf{TP} & \textbf{FP} & \textbf{FN} & \textbf{P(\%)$\uparrow$} & \textbf{R(\%)$\uparrow$} & \textbf{F1(\%)$\uparrow$} & \textbf{MAE$\downarrow$} \\
        \midrule
        Cellpose-SAM & 4773 & 4454 & 319 & 1331 & 93.32 & 76.99 & 84.37 & 18.57 \\ 
        Cellpose-SAMv2 & 4818 & 4488 & 330 & 1297 & 93.15 & 77.58 & 84.66 & 17.52 \\ 
        Cellpose-DINOv3 ViT-B & 5075 & 4643 & 432 & 1142 & 91.49 & 80.26 & 85.51 & 13.82 \\
        Cellpose-DINOv3 ViT-L & 5003 & 4586 & 417 & 1199 & 91.67 & 79.27 & 85.02 & 15.46 \\
        \bottomrule
    \end{tabular}
\end{table}

\subsection{Features extraction}
\label{A.3}
The encoders expect a fixed input size, typically $224\times224$~px, while the nuclei detected by Cellpose-DINOv3 ViT-B vary in size. Each detection must therefore be turned into a fixed-size crop before encoding, and this conversion determines how much of the surrounding tissue reaches the encoder and whether neighboring nuclei appear in the patch. In all configurations the crop is taken in a fixed $64\times64$~px bounding box and upsampled to the encoder resolution.

The configurations below differ only in how the region outside the target nucleus is treated. All start from the same segmentation model output, and the rest of the pipeline is unchanged.

\subsubsection{Neighboring/real background}
\begin{figure}[ht]
    \centering
    \includegraphics[width=1.0\linewidth]{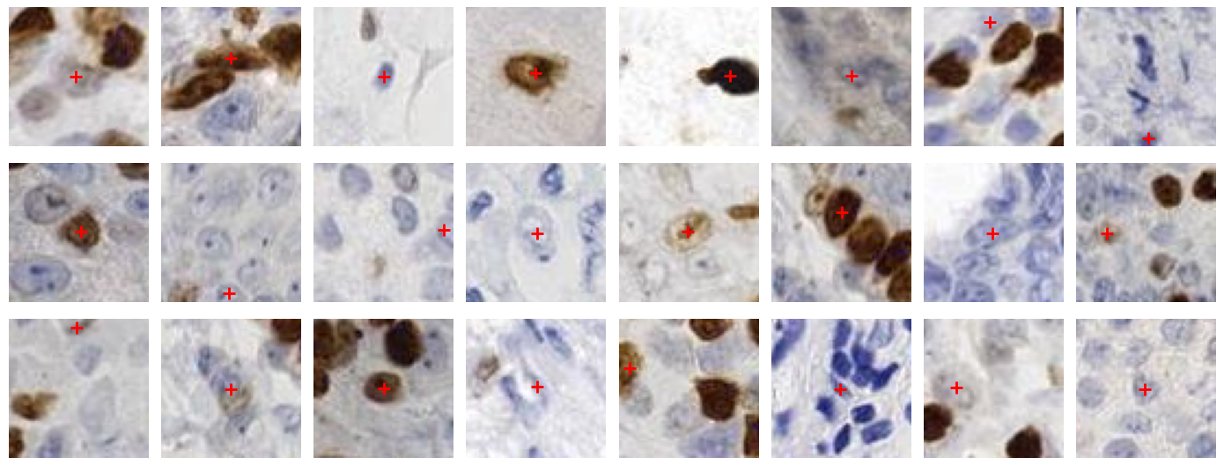}
    \caption{Examples of crops with neighboring or real context background.}
    \label{fig:crop_real}
\end{figure}

In this setting, the crop is cut directly from the original image, without any masking or inpainting of the surroundings. The crop size is fixed to $64\times64$~px, corresponding to the largest cell in the whole dataset plus one pixel of padding, so that every cell fits entirely inside the patch and no rescaling is required. Each crop is centered on the centroid of the target cell. Figure~\ref{fig:crop_real} illustrates examples of crops obtained with this strategy.

This choice keeps the appearance of the patch fully realistic: staining intensity, tissue texture and local context are exactly those seen at inference time, and no artificial boundary is introduced. The counterpart is label ambiguity. Because the crop covers an area large enough for the biggest cell in the dataset, most patches also contain one or several neighbors, which may be larger, more densely stained or more central-looking than the detected cell. The model must therefore learn implicitly that the object to classify is the one at the center of the patch, which is not guaranteed when neighboring cells
belong to a different class. Table~\ref{tab:app2} reports the results for the seven feature encoders under this background strategy, using the masks produced by Cellpose-DINOv3 ViT-B.

\begin{table}[ht]
    \caption{Ki67 cell typing performance on MIST with real context background, after PCA whitening. Precision, recall, F1-score, and MCC are computed over Cellpose-DINOv3 detections matched to ground-truth nuclei, while MAE is computed from the per-tile predicted and ground-truth cell counts.}
    \label{tab:app2}
    \centering
    \footnotesize
    \begin{tabular}{lcccccc} 
        \toprule
        \textbf{Model} & \textbf{Class} & \textbf{Precision(\%)$\uparrow$} & \textbf{Recall(\%)$\uparrow$} & \textbf{F1-Score(\%)$\uparrow$} & \textbf{MAE$\downarrow$} & \textbf{MCC$\uparrow$} \\
        \midrule
        \multirow{2}{*}{EfficientNet-B2} & Ki67+ & 38.80 & 94.34 & 54.98 & 37.79 & \multirow{2}{*}{0.3023} \\
        & Ki67- & 93.19 & 34.24 & 50.08 & 50.43 & \\
        \midrule
        \multirow{2}{*}{ResNet-101} & Ki67+ & 40.63 & 48.67 & 44.29 & 16.52 & \multirow{2}{*}{0.1649} \\
        & Ki67- & 75.14 & 68.57 & 71.71 & 22.02 & \\
        \midrule
        
        \multirow{2}{*}{ConvNeXt V2 Tiny} & Ki67+ & 67.13 & 40.28 & 50.35 & 14.68 & \multirow{2}{*}{0.3757} \\
        & Ki67- & 77.57 & 91.29 & 83.87 & 13.71 & \\
        \midrule
        \multirow{2}{*}{DINOv3 ViT-S/16} & Ki67+ & 44.46 & 97.62 & 61.09 & 31.77 & \multirow{2}{*}{0.4297} \\
        & Ki67- & 97.77 & 46.11 & 62.66 & 44.20 & \\
        \midrule
        
        \multirow{2}{*}{HoVer-NeXt} & Ki67+ & 63.64 & 72.94 & 67.97 & 11.80 & \multirow{2}{*}{0.5265} \\
        & Ki67- & 87.22 & 81.58 & 84.30 & 18.09 & \\
        \midrule
        
        \multirow{2}{*}{UNI2} & Ki67+ & 31.56 & 99.86 & 47.96 & 58.57 & \multirow{2}{*}{0.1119} \\
        & Ki67- & 98.58 & 4.30 & 8.23 & 71.14 & \\
        \midrule
        
        \multirow{2}{*}{Virchow2} & Ki67+ & 43.08 & 99.02 & 60.04 & 34.52 & \multirow{2}{*}{0.4163} \\
        & Ki67- & 98.98 & 42.18 & 59.15 & 47.02 & \\
        \bottomrule
    \end{tabular}
\end{table}

\subsubsection{White background}
\begin{figure}[ht]
    \centering
    \includegraphics[width=1.0\linewidth]{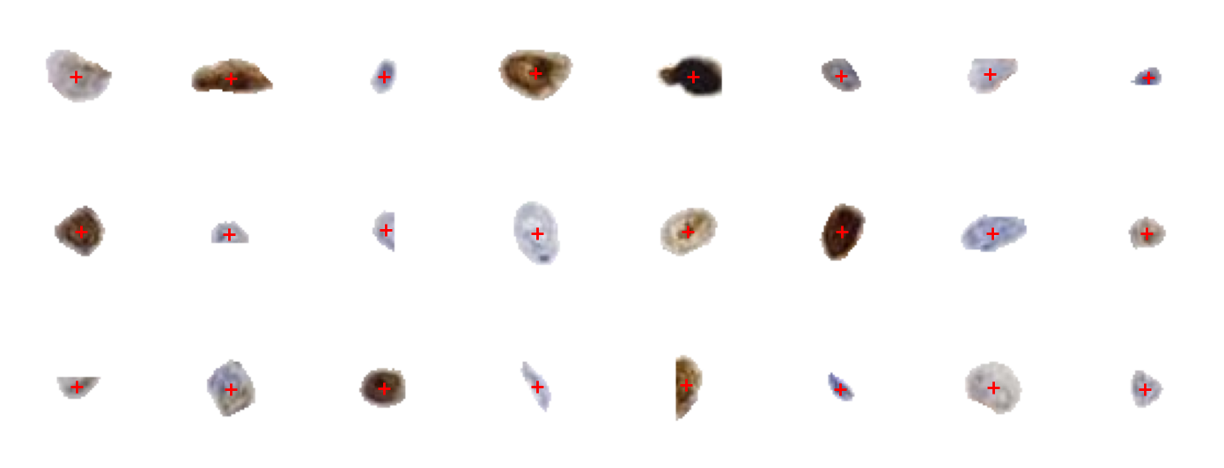}
    \caption{Examples of crops extracted with a white background.}
    \label{fig:crop_white}
\end{figure}

This configuration keeps the fixed $64\times64$~px bounding box of the previous one, but replaces everything that does not belong to the target cell by the white color. Since the surroundings no longer carry any information, the cell is also placed at the center of the bounding box, so that every instance occupies the same position in the patch, Figure~\ref{fig:crop_white} illustrates examples of crops obtained with this strategy.

White maximizes the contrast between the cell and its surroundings. It is also the simplest option to reproduce, since it requires no estimation step. On the other hand it is slightly brighter than the true background of our tiles, whose global median is a light grey, and the resulting edge along the mask boundary is therefore more marked than with the median fill. This is visible in the geometry of the embedding, where the first principal component is associated with nuclear area at between 0.603 and 0.722 for five of the seven encoders, against 0.138 to 0.654 with the median fill. Table~\ref{tab:app3} reports the results for the seven feature encoders under this background strategy, using the masks produced by Cellpose-DINOv3 ViT-B.

\begin{table}[ht]
    \caption{Ki67 cell typing performance on MIST with the white background fill, after PCA whitening. Precision, recall, F1-score, and MCC are computed over Cellpose-DINOv3 detections matched to ground-truth nuclei, while MAE is computed from the per-tile predicted and ground-truth cell counts.}
    \label{tab:app3}
    \centering
    \footnotesize
    \begin{tabular}{lcccccc} 
        \toprule
        \textbf{Model} & \textbf{Class} & \textbf{Precision(\%)$\uparrow$} & \textbf{Recall(\%)$\uparrow$} & \textbf{F1-Score(\%)$\uparrow$} & \textbf{MAE$\downarrow$} & \textbf{MCC$\uparrow$} \\
        \midrule
        \multirow{2}{*}{EfficientNet-B2} & Ki67+ & 95.98 & 93.43 & 94.68 & 5.59 & \multirow{2}{*}{0.9240} \\
        & Ki67- & 97.13 & 98.27 & 97.70 & 11.48 & \\
        \midrule

        \multirow{2}{*}{ResNet-101} & Ki67+ & 94.87 & 94.48 & 94.67 & 5.34 & \multirow{2}{*}{0.9233} \\
        & Ki67- & 97.56 & 97.74 & 97.65 & 12.12 & \\
        \midrule

        \multirow{2}{*}{ConvNeXt V2 Tiny} & Ki67+ & 95.27 & 95.87 & 95.57 & 4.88 & \multirow{2}{*}{0.9361} \\
        & Ki67- & 98.17 & 97.90 & 98.03 & 12.20 & \\
        \midrule

        \multirow{2}{*}{DINOv3 ViT-S/16} & Ki67+ & 95.85 & 93.71 & 94.77 & 5.61 & \multirow{2}{*}{0.9250} \\
        & Ki67- & 97.25 & 98.21 & 97.72 & 11.54 & \\
        \midrule

        \multirow{2}{*}{HoVer-NeXt} & Ki67+ & 94.25 & 95.17 & 94.71 & 5.09 & \multirow{2}{*}{0.9236} \\
        & Ki67- & 97.86 & 97.44 & 97.65 & 12.23 & \\
        \midrule

        \multirow{2}{*}{UNI2} & Ki67+ & 95.70 & 93.36 & 94.51 & 5.66 & \multirow{2}{*}{0.9215} \\
        & Ki67- & 97.10 & 98.15 & 97.62 & 11.45 & \\
        \midrule

        \multirow{2}{*}{Virchow2} & Ki67+ & 95.05 & 92.66 & 93.84 & 5.95 & \multirow{2}{*}{0.9118} \\
        & Ki67- & 96.79 & 97.87 & 97.33 & 11.70 & \\
        \bottomrule
    \end{tabular}
\end{table}

Masking the region outside the target nucleus is necessary rather than merely convenient. With the real background retained, no encoder exceeds 67.97\% Ki67-positive F1-score (Table~\ref{tab:app2}), and the ranking bears no relation to that of the masked configurations, since the crop frequently contains neighbors of the opposite class and nothing identifies which nucleus is the target.

Between the two masked strategies the difference is small. After whitening, all seven encoders fall between 93.8\% and 95.6\% Ki67-positive F1-score under either fill. The white fill is simpler, since it requires no estimation step, and reaches a marginally higher best score at 95.57\% against 94.78\%. We adopt the global median fill for the main results for two reasons. It gives the tighter convergence across encoders after whitening, with a spread of 1.00 F1-score points against 1.73, which is the property of interest in this work. It also leaves the leading directions of the representation less influenced by the mask boundary, since under the white fill the first principal component is associated with nuclear area at between 0.603 and 0.722 for five of the seven encoders, against 0.138 to 0.654 with the median fill.

\subsection{Handcrafted features}
\label{A.4}
We evaluate handcrafted color representations to determine how much of the Ki67 signal can be recovered directly from chromatic information. Features are extracted from the pixels inside each Cellpose-DINOv3 nuclear mask, excluding the surrounding tissue and neighboring nuclei. We consider HED, HSV, and CIE~L$^*$a$^*$b$^*$ independently, as well as a combined representation obtained by concatenating the descriptors from all three color spaces. Each channel is summarized using distributional statistics computed over the nuclear pixels.

For these representations, constant features are removed and the remaining features are standardized before applying PCA, retaining the components that explain 90\% of the variance. We report results without and with PCA whitening, matching the preprocessing applied to the pretrained embeddings. The retained components are clustered globally using $k$-means with $k=2$, and the cluster with the higher mean DAB optical density is assigned to the Ki67-positive class. We also include a simpler DAB baseline, which represents each nucleus only by its mean DAB optical density and separates the two classes using a global Otsu threshold. Since this baseline uses a single channel, its results are unaffected by whitening.

Tables~\ref{tab:app4} and~\ref{tab:app5} report the end-to-end results on MIST and BCData, respectively. On MIST, without whitening, HSV achieves the highest Ki67-positive F1-score among the handcrafted representations at 82.12\%, while the combined HED+HSV+CIE~L$^*$a$^*$b$^*$ representation gives the highest Ki67-negative F1-score at 80.47\%, only marginally above HSV (80.42\%). Whitening improves the F1-score of all color-space sets for both classes, although the Ki67-negative MAE increases for HSV and for the combined representation. The DAB baseline attains high precision for Ki67-positive nuclei but substantially lower recall, showing that mean DAB intensity alone does not capture all of the variation required for robust cell typing.

\begin{table}[ht]
    \centering
    \footnotesize
    \setlength{\tabcolsep}{4pt}
    \caption{End-to-end comparison of color-space feature sets for Ki67-positive and Ki67-negative cell detection and classification on MIST, without and with PCA whitening, using one-to-one Hungarian matching with a $3\,\mu\mathrm{m}$ tolerance.}
    \label{tab:app4}
    \begin{tabular}{l c cccc cccc}
        \toprule
        & & \multicolumn{4}{c}{Ki67-positive cells} & \multicolumn{4}{c}{Ki67-negative cells} \\ 
        \cmidrule(lr){3-6} \cmidrule(lr){7-10}
        \textbf{Methods} & \textbf{Whitening} & \textbf{P(\%)$\uparrow$} & \textbf{R(\%)$\uparrow$} & \textbf{F1(\%)$\uparrow$} & \textbf{MAE$\downarrow$} &
        \textbf{P(\%)$\uparrow$} & \textbf{R(\%)$\uparrow$} & \textbf{F1(\%)$\uparrow$} & \textbf{MAE$\downarrow$} \\
        \midrule
        DAB baseline & No & 96.86 & 61.75 & 75.42 & 10.84 & 80.15 & 77.96 & 79.04 & 9.34 \\
        \midrule
        \multirow{2}{*}{HED} & No & 97.22 & 53.07 & 68.66 & 13.36 & 77.25 & 77.96 & 77.60 & 9.89 \\
         & Yes & 93.99 & 68.43 & 79.20 & 8.54 & 82.69 & 77.45 & 79.99 & 8.86 \\
         \midrule
        \multirow{2}{*}{HSV} & No & 87.66 & 77.23 & 82.12 & 7.50 & 86.12 & 75.42 & 80.42 & 11.39 \\
         & Yes & 88.29 & 80.09 & 83.99 & 5.66 & 88.36 & 76.46 & 81.98 & 11.80 \\
         \midrule
        \multirow{2}{*}{L$^*$a$^*$b$^*$} & No & 84.44 & 71.52 & 77.45 & 9.04 & 83.32 & 74.09 & 78.43 & 11.43 \\
         & Yes & 91.82 & 78.38 & 84.57 & 6.34 & 86.81 & 76.97 & 81.59 & 10.48 \\
         \midrule
        \multirow{2}{*}{HED+HSV+L$^*$a$^*$b$^*$} & No & 94.94 & 70.61 & 80.99 & 7.82 & 83.48 & 77.67 & 80.47 & 9.07 \\
         & Yes & 91.19 & 77.90 & 84.02 & 6.36 & 86.72 & 76.87 & 81.50 & 10.64 \\
        \bottomrule
    \end{tabular}
\end{table}

\begin{table}[ht]
    \centering
    \footnotesize
    \setlength{\tabcolsep}{4pt}
    \caption{End-to-end comparison of color-space feature sets for Ki67-positive and Ki67-negative cell detection and classification on BCData, without and with PCA whitening, using one-to-one Hungarian matching with a $3\,\mu\mathrm{m}$ tolerance.}
    \label{tab:app5}
    \begin{tabular}{l c cccc cccc}
        \toprule
        & & \multicolumn{4}{c}{Ki67-positive cells} & \multicolumn{4}{c}{Ki67-negative cells} \\ 
        \cmidrule(lr){3-6} \cmidrule(lr){7-10}
        \textbf{Methods} & \textbf{Whitening} & \textbf{P(\%)$\uparrow$} & \textbf{R(\%)$\uparrow$} & \textbf{F1(\%)$\uparrow$} & \textbf{MAE$\downarrow$} &
        \textbf{P(\%)$\uparrow$} & \textbf{R(\%)$\uparrow$} & \textbf{F1(\%)$\uparrow$} & \textbf{MAE$\downarrow$} \\
        \midrule
        DAB baseline & No & 82.48 & 70.45 & 75.99 & 2.28 & 77.81 & 80.68 & 79.22 & 4.39 \\
        \midrule
        \multirow{2}{*}{HED} & No & 82.92 & 66.15 & 73.59 & 2.58 & 75.87 & 80.74 & 78.23 & 4.62 \\
         & Yes & 83.05 & 66.41 & 73.80 & 2.56 & 75.93 & 80.74 & 78.26 & 4.61 \\
         \midrule
        \multirow{2}{*}{HSV} & No & 78.33 & 81.94 & 80.10 & 2.08 & 84.13 & 79.40 & 81.70 & 4.05 \\
         & Yes & 52.60 & 63.76 & 57.65 & 4.58 & 73.48 & 63.44 & 68.09 & 6.00 \\
         \midrule
        \multirow{2}{*}{L$^*$a$^*$b$^*$} & No & 77.57 & 83.75 & 80.54 & 2.16 & 85.13 & 78.97 & 81.93 & 4.14 \\
         & Yes & 75.75 & 85.32 & 80.25 & 2.39 & 86.27 & 78.07 & 81.96 & 4.28 \\
         \midrule
        \multirow{2}{*}{HED+HSV+L$^*$a$^*$b$^*$} & No & 80.13 & 78.75 & 79.43 & 1.96 & 82.36 & 80.26 & 81.30 & 4.09 \\
         & Yes & 78.76 & 81.97 & 80.33 & 2.02 & 84.08 & 79.58 & 81.77 & 4.08 \\
        \bottomrule
    \end{tabular}
\end{table}

On BCData, without whitening, CIE~L$^*$a$^*$b$^*$ achieves the highest F1-score for both Ki67-positive and Ki67-negative nuclei, with 80.54\% and 81.93\%, respectively, while HSV performs similarly at 80.10\% and 81.70\%. The combined representation gives the lowest Ki67-positive counting error (MAE of 1.96) but does not surpass the F1-score of CIE~L$^*$a$^*$b$^*$ and HSV. Whitening has little effect on HED and CIE~L$^*$a$^*$b$^*$ and slightly improves the F1-score of the combined representation, which becomes the best whitened representation for Ki67-positive nuclei at 80.33\%. In contrast, whitening substantially degrades HSV, whose Ki67-positive F1-score drops from 80.10\% to 57.65\%.

Since all handcrafted representations share the same nuclear detections, the drop for HSV arises from classification alone. For HSV on BCData, whitening shifts the clustering from the first principal component, which is strongly correlated with DAB optical density ($r=0.94$), to the second, which is dominated by hue statistics and unrelated to DAB ($r=0.03$) but has a multimodal distribution. $k$-means then partly separates the nuclei by hue rather than by Ki67 status. For the other representations, and for HSV on MIST, the clustering remains aligned with the first component after whitening. These results confirm that chromatic information provides a strong signal for Ki67 typing, while the relative effectiveness of different handcrafted representations varies across datasets.

\subsection{Variance structure of the embeddings}
\label{A.5}
\begin{figure}[!ht]
    \centering
    \includegraphics[width=1.0\linewidth]{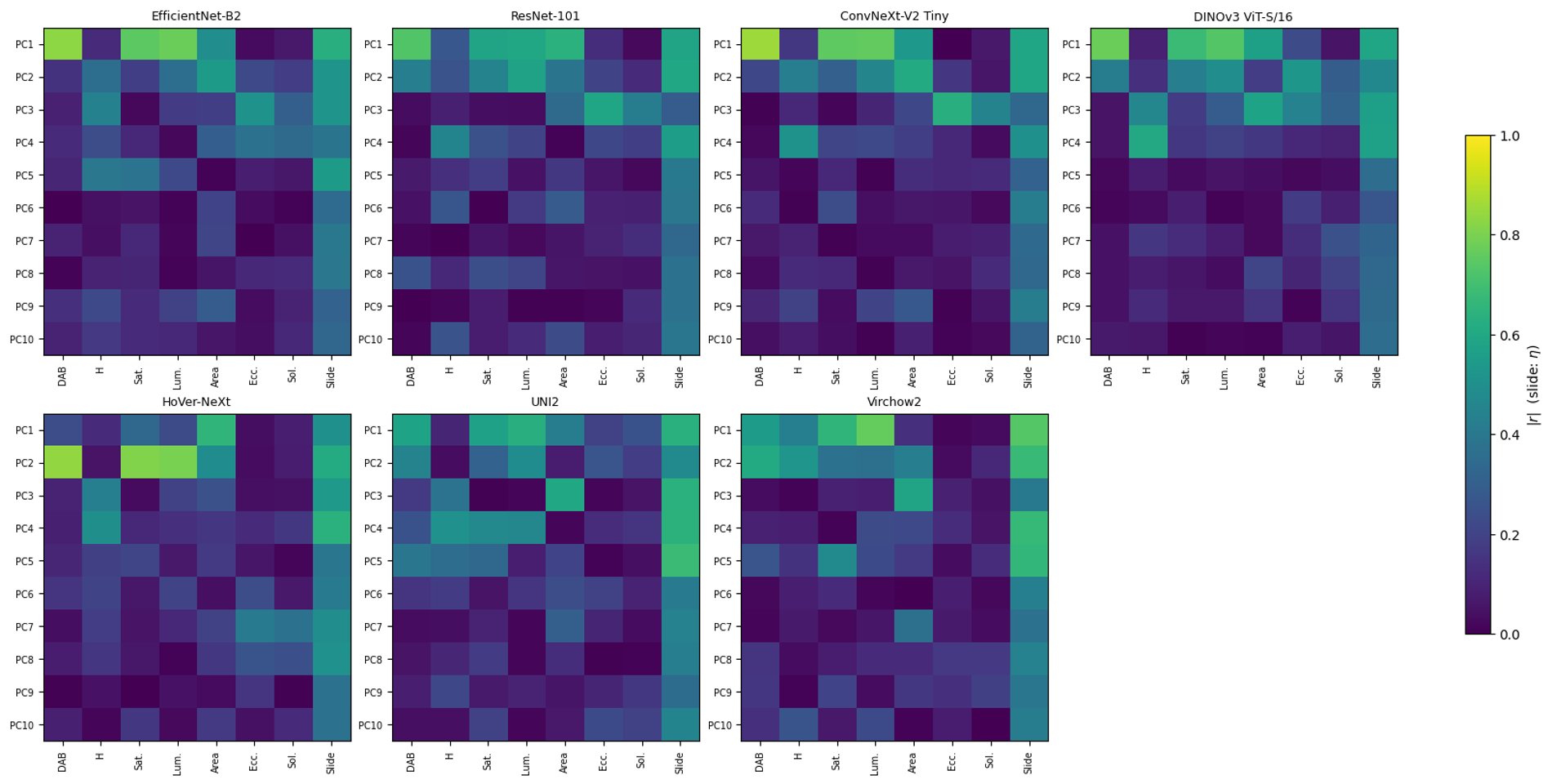}
    \caption{Association between the first ten principal components of each frozen representation and the per-nucleus covariates measured from the image.}
    \label{fig:pc_cov}
\end{figure}

Figure~\ref{fig:pc_cov} reports the association between the first ten principal components of each frozen representation and per-nucleus covariates measured from the image. No annotations are used at any point.

Across encoders the leading components are associated with a mixture of staining, morphology and slide-level covariates, and the structure differs between representations. For EfficientNet-B2, ResNet-101, ConvNeXt-V2 Tiny and DINOv3 ViT-S/16 the DAB association is strongest at the first component and weakens over the components that follow. For HoVer-NeXt it is strongest at the second component, at 0.838 against 0.238 at the first, which is occupied by nuclear area. For UNI2 and Virchow2 slide identity is associated with all ten components, so the dominant variance reflects the acquisition conditions of the tile rather than the appearance of individual nuclei. The stain signal is therefore present in every representation but does not always occupy the directions of largest variance.

Values are absolute Pearson correlations, except for the slide of origin, which is categorical and is reported as the correlation ratio $\eta$.

\subsection{Effect of the whitening exponent}
\label{A.6}
Whitening is applied as a single fixed operation in the main experiments, rescaling every retained component to unit variance. To show that the effect comes from re-balancing the variance structure rather than from information whitening might add, we interpolate between the two extremes by dividing each retained component by its standard deviation raised to the power $\alpha$, so that $\alpha=0$ leaves the components as PCA produces them and $\alpha=1$ gives every component unit variance. Table~\ref{tab:app6} reports the Ki67-positive F1-score across this range.

\begin{table}[ht]
    \caption{Ki67-positive F1-score (\%) as a function of the whitening exponent $\alpha$. The PC column gives the number of components retained at the 90\% variance cutoff.}
    \label{tab:app6}
    \centering
    \footnotesize
    \setlength{\tabcolsep}{3.5pt}
    \begin{tabular}{lcccccccccccc}
        \toprule
        \textbf{Model} & \textbf{PC} & \textbf{0.0} & \textbf{0.1} & \textbf{0.2} & \textbf{0.3} & \textbf{0.4} & \textbf{0.5} & \textbf{0.6} & \textbf{0.7} & \textbf{0.8} & \textbf{0.9} & \textbf{1.0} \\
        \midrule
        EfficientNet-B2  & 114 & 94.52 & 94.66 & 94.66 & 94.68 & 94.76 & 94.79 & 94.85 & 94.85 & 94.84 & 94.98 & 94.67 \\
        ResNet-101       & 201 & 87.81 & 88.51 & 89.53 & 90.04 & 91.52 & 92.40 & 92.65 & 93.31 & 93.60 & 94.11 & 94.40 \\
        ConvNeXt-V2 Tiny &  64 & 92.61 & 92.64 & 92.67 & 93.06 & 93.25 & 93.51 & 93.79 & 93.98 & 94.34 & 94.69 & 94.78 \\
        DINOv3 ViT-S/16  &  54 & 92.81 & 93.07 & 93.29 & 93.63 & 93.66 & 93.99 & 94.24 & 94.63 & 94.77 & 95.12 & 94.58 \\
        HoVer-NeXt       &  16 & 50.37 & 50.45 & 50.63 & 50.60 & 50.83 & 50.63 & 51.87 & 93.62 & 93.67 & 94.25 & 94.19 \\
        UNI2             &  90 & 66.83 & 66.92 & 68.15 & 68.81 & 69.21 & 84.25 & 88.89 & 91.33 & 93.79 & 94.53 & 94.32 \\
        Virchow2         &  38 & 61.44 & 61.58 & 61.60 & 61.45 & 61.63 & 62.35 & 62.59 & 92.77 & 93.45 & 94.30 & 93.80 \\
        \bottomrule
    \end{tabular}
\end{table}

The four encoders whose first component is associated with DAB optical density start close to their final performance and improve gradually. The three remaining encoders behave differently. HoVer-NeXt and Virchow2 stay at their initial values until $\alpha=0.6$ and then gain more than 30 points at the next step ($\alpha=0.7$), while UNI2 improves progressively from $\alpha=0.5$ onward. Partial suppression of the dominant direction therefore leaves the clusters unchanged for as long as that direction still governs the Euclidean distance, and the correct separation appears as soon as it no longer does.

\subsection{Ablation studies on BCData dataset}
\label{A.7}
Tables~\ref{tab:app7} and~\ref{tab:app8} extend the encoder comparison to BCData. As on MIST, the unwhitened representations differ substantially in their ability to recover Ki67 status through clustering. Before whitening, the Ki67-positive F1-score ranges from 52.76\% for HoVer-NeXt to 93.67\% for EfficientNet-B2, corresponding to a spread of 40.91 points. The same pattern is reflected in MCC, which varies from 0.2388 to 0.9034. In particular, HoVer-NeXt, UNI2, and ResNet-101 perform substantially worse than EfficientNet-B2, ConvNeXt-V2 Tiny, and DINOv3 ViT-S/16, despite all models being evaluated with the same segmentation and clustering procedure.

PCA whitening substantially reduces these differences. HoVer-NeXt improves from 52.76\% to 93.89\% Ki67-positive F1-score, UNI2 from 55.74\% to 92.24\%, and ResNet-101 from 73.58\% to 91.46\%. After whitening, all seven encoders fall between 90.89\% and 94.42\% Ki67-positive F1-score, reducing the spread from 40.91 to 3.53 points. The MCC range similarly contracts from 0.665 to 0.052. Ki67-negative performance also becomes substantially more consistent across encoders, with F1-scores between 94.50\% and 96.88\% after whitening. These results reproduce the main trend observed on MIST, although the convergence across encoders is not as tight on BCData.

\begin{table}[!ht]
    \caption{Comparison of pretrained encoders for Ki67 cell typing on BCData \textbf{before} PCA whitening. Classification metrics are computed over Cellpose-DINOv3 detections matched to ground-truth nuclei, while MAE reports per-tile counting error over the predicted cell set.}
    \label{tab:app7}
    \centering
    \footnotesize
    \begin{tabular}{lcccccc} 
        \toprule
        \textbf{Model} & \textbf{Class} & \textbf{Precision(\%)$\uparrow$} & \textbf{Recall(\%)$\uparrow$} & \textbf{F1-Score(\%)$\uparrow$} & \textbf{MAE$\downarrow$} & \textbf{MCC$\uparrow$} \\
        \midrule
        \multirow{2}{*}{EfficientNet-B2} & Ki67+ & 89.51 & 98.23 & 93.67 & 2.73 & \multirow{2}{*}{0.9034} \\
        & Ki67- & 99.02 & 93.96 & 96.42 & 4.43 & \\
        \midrule
        
        \multirow{2}{*}{ResNet-101} & Ki67+ & 63.38 & 87.69 & 73.58 & 5.67 & \multirow{2}{*}{0.5813} \\
        & Ki67- & 91.91 & 73.42 & 81.64 & 6.94 & \\
        \midrule
        
        \multirow{2}{*}{ConvNeXt V2 Tiny} & Ki67+ & 90.51 & 93.87 & 92.16 & 2.13 & \multirow{2}{*}{0.8797} \\
        & Ki67- & 96.72 & 94.84 & 95.77 & 4.32 & \\
        \midrule
        
        \multirow{2}{*}{DINOv3 ViT-S/16} & Ki67+ & 87.97 & 98.20 & 92.80 & 2.93 & \multirow{2}{*}{0.8904} \\
        & Ki67- & 99.00 & 92.95 & 95.88 & 4.59 & \\
        \midrule
        
        \multirow{2}{*}{HoVer-NeXt} & Ki67+ & 47.45 & 59.41 & 52.76 & 7.86 & \multirow{2}{*}{0.2388} \\
        & Ki67- & 75.46 & 65.48 & 70.12 & 7.58 & \\
        \midrule
        
       \multirow{2}{*}{UNI2} & Ki67+ & 45.97 & 70.78 & 55.74 & 8.79 & \multirow{2}{*}{0.2584} \\
       & Ki67- & 78.62 & 56.37 & 65.66 & 8.77 & \\
       \midrule
        
       \multirow{2}{*}{Virchow2} & Ki67+ & 81.40 & 99.65 & 89.60 & 3.77 & \multirow{2}{*}{0.8438} \\
       & Ki67- & 99.79 & 88.06 & 93.56 & 5.18 & \\
       \bottomrule
   \end{tabular}
\end{table}

\begin{table}[t]
    \caption{Comparison of pretrained encoders for Ki67 cell typing on BCData \textbf{after} PCA whitening. Classification metrics are computed over Cellpose-DINOv3 detections matched to ground-truth nuclei, while MAE reports per-tile counting error over the predicted cell set.}
    \label{tab:app8}
    \centering
    \footnotesize
    \begin{tabular}{lcccccc} 
        \toprule
        \textbf{Model} & \textbf{Class} & \textbf{Precision(\%)$\uparrow$} & \textbf{Recall(\%)$\uparrow$} & \textbf{F1-Score(\%)$\uparrow$} & \textbf{MAE$\downarrow$} & \textbf{MCC$\uparrow$} \\
        \midrule
        \multirow{2}{*}{EfficientNet-B2} & Ki67+ & 90.37 & 98.44 & 94.23 & 2.77 & \multirow{2}{*}{0.9121} \\
        & Ki67- & 99.14 & 94.50 & 96.76 & 4.52 & \\
        \midrule
        
        \multirow{2}{*}{ResNet-101} & Ki67+ & 84.85 & 99.18 & 91.46 & 3.34 & \multirow{2}{*}{0.8709} \\
        & Ki67- & 99.53 & 90.71 & 94.92 & 4.92 & \\
        \midrule
        
        \multirow{2}{*}{ConvNeXt V2 Tiny} & Ki67+ & 90.65 & 98.53 & 94.42 & 2.69 & \multirow{2}{*}{0.9150} \\
        & Ki67- & 99.19 & 94.67 & 96.88 & 4.45 & \\
        \midrule
        
        \multirow{2}{*}{DINOv3 ViT-S/16} & Ki67+ & 83.64 & 99.53 & 90.89 & 3.50 & \multirow{2}{*}{0.8629} \\
        & Ki67- & 99.73 & 89.79 & 94.50 & 5.03 & \\
        \midrule
        
        \multirow{2}{*}{HoVer-NeXt} & Ki67+ & 93.46 & 94.32 & 93.89 & 2.53 & \multirow{2}{*}{0.9066} \\
        & Ki67- & 97.00 & 96.54 & 96.77 & 4.23 & \\
        \midrule
        
       \multirow{2}{*}{UNI2} & Ki67+ & 86.32 & 99.03 & 92.24 & 3.20 & \multirow{2}{*}{0.8824} \\
       & Ki67- & 99.45 & 91.76 & 95.45 & 4.80 & \\
       \midrule
        
       \multirow{2}{*}{Virchow2} & Ki67+ & 84.15 & 99.44 & 91.16 & 3.45 & \multirow{2}{*}{0.8667} \\
       & Ki67- & 99.68 & 90.17 & 94.69 & 4.97 & \\
       \bottomrule
   \end{tabular}
\end{table}

The representation-level analysis reproduces the pattern observed on MIST. Despite substantial variation in unwhitened clustering performance, all seven encoders strongly encode Ki67 status, with supervised probe AUCs between 0.9945 (ResNet-101) and 0.9960 (Virchow2). Unwhitened performance again follows the variance structure of the embeddings, the association between PC1 and DAB optical density correlates with Ki67-positive F1-score at \(r=0.999\). HoVer-NeXt and UNI2 show weak PC1--DAB associations but recover the DAB signal more strongly in PC2, whereas the encoders with high unwhitened performance show stronger DAB association in PC1 (Table~\ref{tab:app9}). Whitening reduces the Ki67-positive F1-score spread from 40.91 to 3.53 points.

\begin{table}[ht]
    \caption{Representation analysis on BCData. Probe AUC measures linear decodability of Ki67 status. DAB associations are absolute Pearson correlations with the corresponding principal component.}
    \label{tab:app9}
    \centering
    \small
    \begin{tabular}{lccc}
    \toprule
    \textbf{Model} & \textbf{Probe AUC} & \textbf{PC1--DAB} & \textbf{Strongest DAB PC} \\
    \midrule
    EfficientNet-B2      & 0.9958 & 0.861 & PC1 (0.861) \\
    ResNet-101           & 0.9945 & 0.597 & PC1 (0.597) \\
    ConvNeXt-V2 Tiny     & 0.9949 & 0.854 & PC1 (0.854) \\
    DINOv3 ViT-S/16      & 0.9956 & 0.870 & PC1 (0.870) \\
    HoVer-NeXt           & 0.9957 & 0.256 & PC2 (0.777) \\
    UNI2                  & 0.9959 & 0.297 & PC2 (0.672) \\
    Virchow2              & 0.9960 & 0.809 & PC1 (0.809) \\
    \bottomrule
    \end{tabular}
\end{table}

% \subsection{Variability across images}
% \label{A.8}
% To assess whether the differences between K-CLUE and the whitened HED+HSV+L$^*$a$^*$b$^*$ baseline are meaningful, we estimate 95\% confidence intervals for the pooled F1 difference using a paired bootstrap over images ([B] resamples), averaged over five runs. On MIST, K-CLUE is slightly better for both classes, but the intervals include zero. On BCData, the handcrafted baseline is slightly but consistently better for both classes, by less than one F1 point. Variation across the five runs is below 0.1 F1 points.

% \begin{table}[ht]
%     \centering
%     \footnotesize
%     \caption{Pooled F1-scores (\%) of K-CLUE and the whitened HED+HSV+L$^*$a$^*$b$^*$ baseline, with the difference (K-CLUE $-$ handcrafted) and its 95\% confidence interval from a paired bootstrap over images, averaged over five runs.}
%     \label{tab:app_bootstrap}
%     \begin{tabular}{ll cccc}
%         \toprule
%         \textbf{Dataset} & \textbf{Class} & \textbf{K-CLUE} & \textbf{Handcrafted} & \textbf{$\Delta$} & \textbf{95\% CI} \\
%         \midrule
%         \multirow{2}{*}{MIST} & Ki67+ & 85.02 & 84.02 & $+1.00$ & $[-0.03,\ 2.13]$ \\
%          & Ki67$-$ & 81.92 & 81.50 & $+0.41$ & $[-0.02,\ 0.80]$ \\
%         \midrule
%         \multirow{2}{*}{BCData} & Ki67+ & 79.61 & 80.37 & $-0.76$ & $[-1.45,\ -0.08]$ \\
%          & Ki67$-$ & 81.16 & 81.81 & $-0.65$ & $[-1.04,\ -0.27]$ \\
%         \bottomrule
%     \end{tabular}
% \end{table}

\end{document}